\documentclass{article}

\usepackage[numbers]{natbib}
\usepackage[final]{neurips_data_2024}

\usepackage[utf8]{inputenc} % allow utf-8 input
\usepackage[T1]{fontenc}    % use 8-bit T1 fonts
\usepackage{hyperref}       % hyperlinks
\usepackage{url}            % simple URL typesetting
\usepackage{booktabs}       % professional-quality tables
\usepackage{amsfonts}       % blackboard math symbols
\usepackage{nicefrac}       % compact symbols for 1/2, etc.
\usepackage{microtype}      % microtypography
\usepackage{xcolor}         % colors

\usepackage{amsmath}
\usepackage{dsfont}
\usepackage{bbm}
\usepackage[noend]{algpseudocode}
\usepackage{algorithm}
\usepackage{natbib}
\usepackage{pdfpages}
\usepackage{url}

\usepackage{amsthm}
\newtheorem*{proof*}{Proof} 

\usepackage{multirow}
\usepackage{multicol}
\usepackage{graphicx}
\usepackage{wrapfig}
\usepackage[normalem]{ulem}
\useunder{\uline}{\ul}{}
\usepackage{tcolorbox}

\usepackage{amsmath}

\usepackage{caption}
\usepackage{array}

\usepackage{varwidth}
\usepackage{bbding}
\usepackage{pifont}
\usepackage{wrapfig}

\usepackage{subcaption}
\usepackage{xspace}

\newcommand{\benchmarkname}{ProGraph\xspace}
\newcommand{\datasetname}{LLM4Graph\xspace}

\title{Can Large Language Models Analyze Graphs like Professionals? A Benchmark, Datasets and Models}

\author{Xin Li$^{1}\thanks{Equal Contribution.}$, Weize Chen$^{2*}$, Qizhi Chu$^{1}$, Haopeng Li$^{3}$, Zhaojun Sun$^3$, Ran Li$^2$, Chen Qian$^2$\\\textbf{ Yiwei Wei$^4$, Zhiyuan Liu$^{2,5}$, Chuan Shi$^1$, Maosong Sun$^{2,5}$, Cheng Yang$^1\thanks{Corresponding author.}$}\\
\\
$^1$ School of Computer Science, Beijing University of Posts and Telecommunications,\\
$^2$ Department of Computer Science and Technology, Tsinghua University,\\
$^3$ School of Artificial Intelligence, Beijing University of Posts and Telecommunications,\\
$^4$ College of Petroleum Engineering, China University of Petroleum (Beijing) at Karamay,\\
$^5$ Institute for Artificial Intelligence, Tsinghua University\\
\\
\texttt{lixin4sky@bupt.edu.cn},\quad \texttt{chenwz21@mails.tsinghua.edu.cn}, \quad
\texttt{yangcheng@bupt.edu.cn} \\
}

\begin{document}

\maketitle

\begin{abstract}

The need to analyze graphs is ubiquitous across various fields, from social networks to biological research and recommendation systems. Therefore, enabling the ability of large language models (LLMs) to process graphs is an important step toward more advanced general intelligence. However, current LLM benchmarks on graph analysis require models to directly reason over the prompts describing graph topology, and are thus limited to small graphs with only a few dozens of nodes. In contrast, human experts typically write programs based on popular libraries for task solving, and can thus handle graphs with different scales. To this end, a question naturally arises: \textit{can LLMs analyze graphs like professionals}? In this paper, we introduce \benchmarkname, a manually crafted benchmark containing 3 categories of graph tasks. The benchmark expects solutions based on programming instead of directly reasoning over raw inputs. Our findings reveal that the performance of current LLMs is unsatisfactory, with the best model achieving only 36\% accuracy. To bridge this gap, we propose \datasetname datasets, which include crawled documents and auto-generated codes based on 6 widely used graph libraries. By augmenting closed-source LLMs with document retrieval and fine-tuning open-source ones on the codes, we show 11-32\% absolute improvements in their accuracies. Our results underscore that the capabilities of LLMs in handling structured data are still under-explored, and show the effectiveness of \datasetname in enhancing LLMs' proficiency of graph analysis. The benchmark, datasets and enhanced open-source models are available at \url{https://github.com/BUPT-GAMMA/ProGraph}.

\end{abstract}

\section{Introduction}

\textbf{Background.} Large language models (LLMs)~\cite{brown2020language, geminiteam2024gemini, AnthropicClaude3} are parameter-rich neural networks trained on a vast amount of text data to understand and generate human language. LLMs can not only handle classical natural language processing tasks like translation, but also benefit task solving in various domains such as code generation~\cite{qian2023communicative}, logical reasoning~\cite{pan-etal-2023-logic}, and mathematical calculation~\cite{RomeraParedes2023MathematicalDF}.

\textbf{Previous LLM Benchmarks for Graph Analysis.} Recently, many researchers have proposed extending LLMs to scenarios that require graph understanding and analysis~\cite{huang2023promptbased, tang2024graphgpt}. As graph is a very commonly used data structure in real-world services (\textit{e.g.,} social networks~\cite{tang2010graph, myers14information, viegas2004social} and urban computing~\cite{li2024urbangpt, zou2024deep, yan2024urbanclip}),  enabling the ability of LLMs to process graphs is an important step toward more advanced general intelligence. To this end, several benchmarks have been developed to evaluate such ability. For example, NLGraph~\cite{NEURIPS2023_622afc4} and GraphInstruct~\cite{luo2024graphinstruct} investigate whether LLMs can understand and compute basic graph properties, such as counting node degrees or finding the shortest path for a node pair. GraphTMI~\cite{das2024modality} and GPT4Graph~\cite{guo2023gpt4graph} also consider typical learning tasks such as node classification. LLM4DyG~\cite{zhang2024llm4dyg} further extends the tasks to dynamic graphs.

\textbf{Limitations of Existing Benchmarks.} However, from the perspective of practicality, we argue that previous benchmarks have three major drawbacks. Firstly, the problems in these work require LLMs to read through the adjacency lists of graphs from prompts before answering specific questions. Consequently, the graph sizes in their benchmarks are rather small (typically with a few dozens of nodes), due to the length limitation of LLMs. Being able to compute the shortest path on a small graph does not mean that the same can be done on a real graph with millions of nodes. Secondly, the desired solving process in these work requires step-by-step reasoning fully based on LLMs. But even with the help of Chain-of-Thought (CoT)~\cite{Wei2022ChainOT, dai2024imitation}, the reasoning depths of current LLMs are still shallow~\cite{li2024faithful, lam2024causalchaos}. Consequently, LLMs might be able to count triangles one by one in a small graph with 10 nodes, and will inevitably fail for large graphs. Thirdly, the problem descriptions in these work are abstract and monotonous in form, lacking context from real-world application scenarios.

\textbf{Inspirations from Human Experts.} Consider the scenario that a human expert is asked to find the shortest path between two nodes in a million-scale graph, she will probably write a few lines of Python codes based on NetworkX\cite{networkx}, instead of directly reasoning over the raw inputs. To this end, a question naturally arises: \textit{can LLMs analyze graphs like professionals}? Fortunately, most popular LLMs have shown the ability to write codes and utilize various application programming inferfaces (APIs), making it possible to analyze graphs via API calling as human experts will do. Compared with direct reasoning in previous benchmarks, generating a few lines of codes requires much shallower reasoning depths for LLMs, but can solve more complex problems.

\textbf{Benchmark.} In this paper, we propose \benchmarkname benchmark to evaluate the capability of LLMs in leveraging external APIs for graph analysis. The benchmark includes 512 problems with hand-crafted questions and answers (QA pairs). The problems cover three categories of tasks: basic graph theory, graph statistical learning, and graph embedding, and can be solved based on six popular Python libraries. In the questions, graphs can be either described by natural language or stored in files, and thus can scale to $10^6$ in our benchmark. To improve the diversity of problem descriptions and align with real-world scenarios, the questions are rephrased in a role-play manner based on GPT-4 turbo~\cite{openai2024gpt4}. In the answers, we provide reference code, key APIs and execution results. We also design an automated evaluation process that aligns well with human judgement.

\begin{table}[t]
\centering
\caption{Comparisons among different graph analysis benchmarks for LLMs.}
\label{tab:benchcom}
\resizebox{\textwidth}{!}{
\begin{tabular}{@{}lccccccc@{}}
\toprule
\textbf{Aspects} & \begin{tabular}[c]{@{}c@{}}\textbf{\benchmarkname}  \\ (this work)   \end{tabular}
         & \begin{tabular}[c]{@{}c@{}}\textbf{NLGraph}       \\ (\cite{NEURIPS2023_622afc4})       \end{tabular}
         & \begin{tabular}[c]{@{}c@{}}\textbf{LLM4DyG}       \\ (\cite{zhang2024llm4dyg})       \end{tabular}
         & \begin{tabular}[c]{@{}c@{}}\textbf{GraphTMI}      \\ (\cite{das2024modality})       \end{tabular}
         & \begin{tabular}[c]{@{}c@{}}\textbf{GraphInstruct} \\ (\cite{luo2024graphinstruct})       \end{tabular}
         & \begin{tabular}[c]{@{}c@{}}\textbf{GPT4Graph}     \\ (\cite{guo2023gpt4graph})       \end{tabular} 
         & \begin{tabular}[c]{@{}c@{}}\textbf{GraphWiz}     \\ (\cite{graphwiz})       \end{tabular} \\
\midrule
Basic Graph Theory  & \textcolor{green}{\ding{51}}
                    & \textcolor{green}{\ding{51}}
                    & \textcolor{green}{\ding{51}}
                    & \textcolor{green}{\ding{51}}
                    & \textcolor{green}{\ding{51}}
                    & \textcolor{green}{\ding{51}} 
                    & \textcolor{green}{\ding{51}} \\
Graph Statistical Learning     & \textcolor{green}{\ding{51}} & \textcolor{red}{\ding{55}} & \textcolor{red}{\ding{55}} & \textcolor{green}{\ding{51}} & \textcolor{red}{\ding{55}} & \textcolor{green}{\ding{51}} &\textcolor{red}{\ding{55}} \\
Graph Embedding  & \textcolor{green}{\ding{51}} & \textcolor{green}{\ding{51}} & \textcolor{red}{\ding{55}} & \textcolor{red}{\ding{55}} & \textcolor{red}{\ding{55}} & \textcolor{red}{\ding{55}} & \textcolor{red}{\ding{55}}\\
Access to External APIs     & \textcolor{green}{\ding{51}} & \textcolor{red}{\ding{55}} & \textcolor{red}{\ding{55}} & \textcolor{red}{\ding{55}} & \textcolor{red}{\ding{55}} & \textcolor{red}{\ding{55}} & \textcolor{red}{\ding{55}} \\
Real-world Context     & \textcolor{green}{\ding{51}} & \textcolor{red}{\ding{55}} & \textcolor{red}{\ding{55}} & \textcolor{red}{\ding{55}} & \textcolor{red}{\ding{55}} & \textcolor{red}{\ding{55}} & \textcolor{red}{\ding{55}} \\
 Scalability & up to ${10^6}$ & up to $10^1$ & up to $10^1$ & up to $10^2$ & up to $10^1$ & up to $10^1$ & up to $10^2$ \\
\bottomrule
\end{tabular}
}
\vspace{-2em}
\end{table}

\textbf{Datasets and Models.} To facilitate LLMs to solve these problems via programming, we construct the \datasetname datasets with both document and code data. The document dataset contains API information crawled from the official documents of the six Python libraries. The code dataset includes 29,260 QA pairs automatically generated by back-instructing~\cite{Wang2022SelfInstructAL} GPT-4 turbo. To enable CoT learning, we further introduce the thought on the document information of relevant APIs to the answers of the code dataset as prefixes. To demonstrate the value of our datasets, we enhance closed-source LLMs by extracting relevant information from the document dataset as RAG (retrieval-augmented generation), and improve open-source ones by instruction tuning over the code dataset. Besides the datasets, the improved open-source LLMs are also released for future researches.

\textbf{Key Results.} The accuracies of closed-source models (Claude, GPT and Gemini) on \benchmarkname are 25-36\%, and can be improved to 37-46\% with RAG using \datasetname as the retrieval pool. The accuracies of open-source models (Llama3~\cite{llama3modelcard} and Deepseek Coder~\cite{deepseek-coder}) are only 12-24\%, but can be improved to 45-47\% through instruction-tuning on \datasetname. These results show that \benchmarkname is challenging for current LLMs, and \datasetname can significantly enhance their performance.

\textbf{Contributions.} (1) To the best of our knowledge, we are the first work exploring the ability of LLMs to analyze graphs with external APIs. The utilization of external APIs is practical and powerful in real-world scenarios.
(2) To evaluate such ability, we propose a novel and challenging \benchmarkname benchmark with hand-crafted QA pairs covering three categories of tasks, \textit{i.e.,} basic graph theory, graph statistical learning, and graph embedding. 
(3) We develop \datasetname datasets containing both crawled document data and auto-generated code data. Experimental results demonstrate that our datasets can substantially enhance the performance of both closed-source and open-source LLMs. The improved open-source models are released together with the datasets for future researches.

\section{Related Work}
\label{sec:related-work}

\textbf{LLM for Graphs.}
Recent efforts leverage the strong generalization ability of LLMs for graph understanding and analysis. Benchmarks like NLGraph~\citep{NEURIPS2023_622afc4} and GraphInstruct~\citep{luo2024graphinstruct} evaluate LLMs' graph reasoning abilities, finding that while LLMs have some capabilities, they are not very strong. GraphTMI~\cite{das2024modality} assesses LLMs' performance on tasks like node classification and link prediction using different input formats. GPT4Graph~\cite{guo2023gpt4graph} evaluates LLMs' understanding of graph structure data in various formats, and LLM4DyG~\cite{zhang2024llm4dyg} studies LLMs in dynamic graphs, introducing techniques to improve performance. GraphWiz~\cite{graphwiz} employs two approaches to optimize reasoning paths for solving basic graph problems. Another approach combines LLMs with graph neural networks (GNNs) to enhance learning tasks, leveraging text attribute processing or direct graph task handling through techniques like prompt learning and instruction tuning~\cite{huang2023promptbased, duan2023simteg, chen2024labelfree, xia2024opengraph, tang2024graphgpt, he2024unigraph, tang2024higpt, GraphTranslator, chai2023graphllm}. However, these works are often specialized for classification tasks and do not handle complex graph tasks.

\textbf{LLM Benchmarks.} LLMs perform strongly in text processing, mathematical computation, and code generation. Text processing benchmarks evaluate capabilities in machine translation, summarization, and comprehension~\cite{amrhein-aces-2022, moghe2024machine, xu2023superclue, luo2024factual, Dua2019DROPAR, Joshi2017TriviaQAAL}. Mathematical computation benchmarks assess understanding of numbers and elementary math concepts~\cite{cobbe2021training, taylor2022galactica}. Specialized field benchmarks, such as those in biology, chemistry, and finance, evaluate LLMs' expertise in specific domains~\cite{Sarwal2023BioLLMBenchAC, Guo2023WhatCL, xie2024finben}. Code generation benchmarks like HumanEval and MBPP focus on function-oriented tasks~\cite{chen2021evaluating, austin2021program}, while our proposed \benchmarkname considers task-oriented code generation.

\textbf{Enhancement Techniques for LLMs.}
There are many techniques to enhance the performance of LLMs~\cite{ouyang2022training, wang2023selfinstruct}. Among these, two important techniques are Chain-of-Thought (CoT)~\cite{Wei2022ChainOT} and retrieval-augmented generation (RAG)~\cite{REALM}. CoT allows the model to mimic human thinking by reasoning step-by-step rather than directly providing an answer, significantly enhancing the logical analysis and reasoning capabilities of LLMs. RAG reduces the LLMs' hallucinations by allowing them to access relevant information before generating answers, improving LLMs' accuracy and reliability across various tasks.
%, such as instruction tuning and in-context learning

\section{Benchmark}

\begin{figure}
    \centering
    \includegraphics[width=\textwidth]{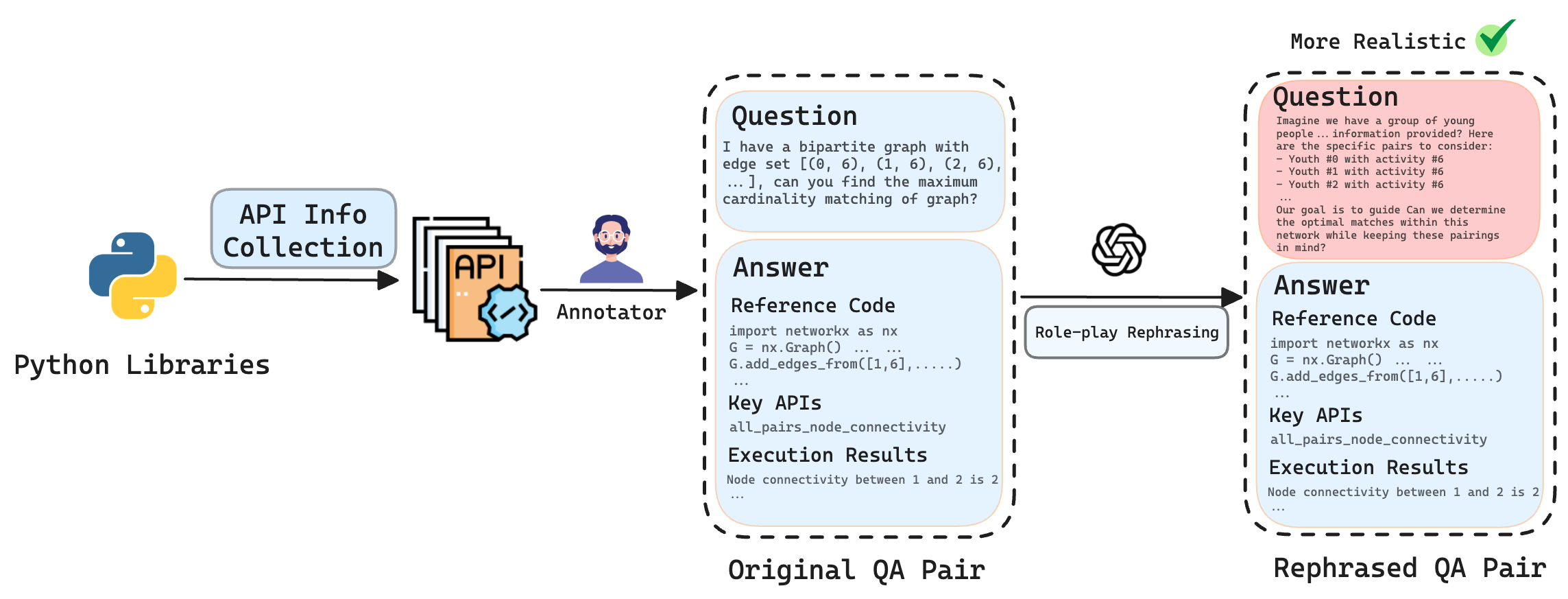}
    \caption{\textbf{The pipeline of \benchmarkname benchmark construction.} We first collect the API documents of 6 frequently-used graph reasoning Python libraries. Then human  annotators read the API information, use random graph generators to generate graph data, write questions based on API information, and give the results and the corresponding python codes, forming the Original QA pairs. Finally, GPT-4 rephrases abstract questions and integrates them into real-world scenarios.}
    \vspace{-2em}
    \label{fig:pipeline-benchmark}
\end{figure}

To evaluate the ability of LLMs in graph analysis, we introduce the \benchmarkname benchmark, featuring 512 problems in three categories. These hand-crafted problems include questions and answers with two difficulty levels. To enhance diversity and realism, we leverage GPT-4 turbo to rephrase the questions in a role-playing manner, followed by manual verification for correctness. Finally, given the high consistency of answer judgments between humans and GPT-4o, we employ GPT-4o to automate the evaluation. We compare the proposed benchmark with previous ones in Table~\ref{tab:benchcom}, present more discussions about related work in Appendix \ref{sec:related-work}, and show a benchmark example in Appendix~\ref{sec:prograph-benchmark-example}.

\subsection{Tasks}

The proposed \benchmarkname benchmark considers three categories of tasks:

\textbf{Category 1: Basic Graph Theory.} Graph theory primarily studies the fundamental concepts of graphs, including types, properties, classical algorithms and many other basic operations. For example, some problems in this category aim to check whether a graph is acyclic, compute the centrality of nodes, or find the maximum cardinality matching.

\textbf{Category 2: Graph Statistical Learning.} Graph statistical learning utilizes a probabilistic model to extract useful information from nodes, edges, or the entire topology for various tasks. In this work, we mainly focus on graph clustering and sampling techniques. For example, detecting communities in a graph with Louvain algorithm~\cite{Blondel2008FastUO}, or sampling a subgraph based on random walk.

\textbf{Category 3: Graph Embedding.} Graph embedding technique aims to learn real-valued vectors for nodes in a graph, where similar nodes are expected to have similar vectors. The learned vectors can be used as features to enhance the performance of downstream tasks. For example, a typical problem in this category will require the learned embeddings of DeepWalk algorithm~\cite{Perozzi2014DeepWalkOL} for a given graph.

\begin{table}[h]
  \centering
  \vspace{-1.5em}
  \caption{Statistics of \benchmarkname.}
  \label{tab:combined}
  \resizebox{0.85\textwidth}{!}{
  \begin{tabular}{@{}lcccccc@{}}
    \toprule
    & \multicolumn{4}{c}{\textbf{Question Type}} & \multicolumn{2}{c}{\textbf{Answer Difficulty}} \\
    \cmidrule(lr){2-5}\cmidrule(lr){6-7}
    % \midrule
    & True/False & Calculation & Drawing & Hybrid & Easy & Hard \\
    \midrule
    Basic Graph Theory & 32 & 240 & 25 & 15 & 257 & 55 \\
    Graph Statistical Learning & 7 & 115 & 7 & 25 & 43 & 111 \\
    Graph Embedding & 0 & 30 & 0 & 16 & 0 & 46 \\
    \midrule
    Total  & 39 & 385 & 32 & 56 & 300 & 212 \\
    \bottomrule
  \end{tabular}
  }
  \vspace{-1em}
\end{table}

\begin{comment}
\begin{table}[h]
  \centering
  \caption{Different Levels}
  \label{tab:level}
  \begin{tabular}{@{}lcc@{}}
    \toprule
    & Level 1 & Level 2 \\
    \midrule
    Task 1 & 257 & 55 \\
    Task 2 & 43 & 111 \\
    Task 3 & 0 & 46 \\
    \midrule
    Total  & 300 & 212 \\
    \bottomrule
  \end{tabular}
\end{table}

\begin{table}[h]
  \centering
  \caption{Different answer difficulties}
  \label{tab:type}
  \begin{tabular}{@{}lcccc@{}}
    \toprule
    & True/False & Calculation & Drawing & Hybrid \\
    \midrule
    Task 1 & 32 & 240 & 25 & 15 \\
    Task 2 & 7 & 115 & 7 & 25 \\
    Task 3 & 0 & 30 & 0 & 16 \\
    \midrule
    Total  & 39 & 385 & 32 & 56 \\
    \bottomrule
  \end{tabular}
\end{table}
\end{comment}

\subsection{Human Annotation}

To create high-quality problems for benchmarking, we invite human annotators to develop questions and answers based on the following guidelines, and the annotation manual is in Appendix \ref{annotation-manual}.

\textbf{Python Libraries.} In this work, we consider six popular libraries to support the above three task categories, \textit{i.e.,} NetworkX~\cite{networkx} and igraph~\cite{igraph} for basic graph theory, CDlib~\cite{CDlib}, graspologic~\cite{graspologic} and Little Ball of Fur~\cite{littleballoffur} for graph statistical learning, and Karate Club~\cite{karateclub} for graph embedding. 

\textbf{Question Design.} First of all, human annotators need to either manually design or use some random graph generators to obtain a graph. Then, based on the API documents of the six libraries, the annotators are asked to design questions with one of the four types: true/false, calculation, drawing, and hybrid questions. Here hybrid questions are composed of two or more of the first three types. Here we present a calculation question from the basic graph theory category as an example:

\begin{minipage}[t]{0.48\textwidth}
\begin{tcolorbox}[colback=gray!10, colframe=black, rounded corners, boxrule=1.5pt, fontupper=\normalsize, left=2mm, right=2mm, top=1mm, bottom=1mm]
    \textbf{Question} \newline
    I have a graph with an edge set [(1, 2), \newline
    (1, 3), (2, 3), (2, 4), (3, 5), (4, 5)], \newline
    can you help me compute node connectivity\newline
    for all pairs of nodes and print the node\newline
    connectivity for each pair?
\end{tcolorbox}
\end{minipage}
\hfill
\begin{minipage}[t]{0.48\textwidth}
\begin{tcolorbox}[colback=gray!10, colframe=black, rounded corners, boxrule=1.5pt, fontupper=\normalsize, left=2mm, right=2mm, top=1mm, bottom=1mm]
    \textbf{Reference Code} \newline
    ** Python Code ** \newline
    \textbf{Key APIs} \newline
    all\_pairs\_node\_connectivity \newline
    \textbf{Execution Results} \newline
    Node connectivity between 1 and 2 is 2 \newline
    ...
\end{tcolorbox}
\end{minipage}

\textbf{Answer Construction.} Based on the proposed question, human annotators need to further provide the code, the number of involved APIs, and the execution result. Based on the number of APIs, we categorize the problems into two difficulty levels: easy level involves one API, while the hard level involves multiple APIs. Here's an example of the collected data for the above question:

\subsection{Role-Play Rephrasing}
\label{role-play-rephrasing}

To make questions more realistic, we rephrase them in a role-play manner. First, GPT-4 turbo generates hundreds of job titles and descriptions. Then, we randomly select a job for a given problem. Using GPT-4 turbo, we mimic the profession's characteristics, knowledge, and style to integrate the question into a real-world context. We manually review the modified questions to ensure they maintain the same semantics and graph structures as the original ones. A complete example of rephrasing is in Appendix \ref{sec:role-play-rephrasing-example}, and here we present the rephrased question of the above problem: 

\begin{tcolorbox}[colback=gray!10, colframe=black, rounded corners, boxrule=1.5pt, fontupper=\normalsize, left=2mm, right=2mm, top=1mm, bottom=1mm]

    \textbf{Rephrased Question} \newline
We're examining a simplified model of an ecosystem where [...], we've mapped out a series of interactions as follows: [(1, 2), (1, 3), (2, 3), (2, 4), (3, 5), (4, 5)]. [...] Can we analyze our network to reveal the minimum number of species that would need to be removed to disrupt the direct connection between any two species in this web? [...]
\end{tcolorbox}

\subsection{Automated Evaluation}

\textbf{Metrics.} To evaluate the ability of LLMs to solve these problems, we consider two metrics: pass rate and accuracy. Pass rate measures the ratio of executable code generated by an LLM, while accuracy measures the ratio of correct answers from the executable code. Accuracy is always no higher than the pass rate and is considered the more important metric as it evaluates final performance.

\textbf{Process.} Evaluating diverse answer formats with rule-based matching is challenging, and human evaluation is too labor-intensive. Thus, we automate evaluation using GPT-4o. First, we extract code snippets from LLM-generated answers using regular expressions. GPT-4o is then asked to check the correctness given the execution result. For problems with certain answers, such as true/false or calculation questions, GPT-4o assigns 1 point if the execution result matches the reference code's result, and 0 otherwise. For other problems, like drawing questions, GPT-4o matches key API usage: if the generated code contains $m$ out of $n$ key APIs, the accuracy point is $m/n$.

\begin{wraptable}{r}{0.33\textwidth}
% \begin{table}[h]
  \vspace{-0em}
  \centering
  \caption{Self-Consistency (SC) and Human-Consistency (HC) of automated evaluation.}
  \vspace{-0.8em}
  \label{tab: GPTEval}
  \begin{tabular}{@{}lcc@{}}
    \toprule
    & \textbf{SC} (\%) & \textbf{HC} (\%) \\
    \midrule
    Category 1 & 98.9 & 97.3 \\
    Category 2 & 98.9 & 96.4 \\
    Category 3 & 98.6 & 97.8 \\
    \midrule
    Overall & 98.9 & 97.5 \\
    \bottomrule
  \end{tabular}
% \end{table}
  \vspace{-1em}
\end{wraptable}

\textbf{Rationale.} To validate GPT-based evaluation, we measure its stability (self-consistency) and alignment with human judgments (human-consistency). Higher stability means judgment scores are consistent across multiple evaluations, while higher human alignment indicates better quality. We use the agreement metric~\citep{NEURIPS2023_91f18a1} to assess these consistencies. 
For $n$ evaluations of the same answer, we take the highest number of evaluations $m$ that received the same score and divide it by $n$ to get the consistency. 
Self-consistency is the agreement among three GPT-4o evaluations, and human-consistency is the agreement between one GPT-4o evaluation and a manual evaluation. We evaluate all 512 problems with answers from \textit{Claude 3 Opus RAG 7}, the best-performing closed-source model which will be introduced in Section \ref{sec:datasets-and-models-models}, and present the results in Table \ref{tab: GPTEval}, showing high self-consistency and human-consistency.

\section{Datasets and Models}

\begin{figure}
    \centering
    \includegraphics[width=\textwidth]{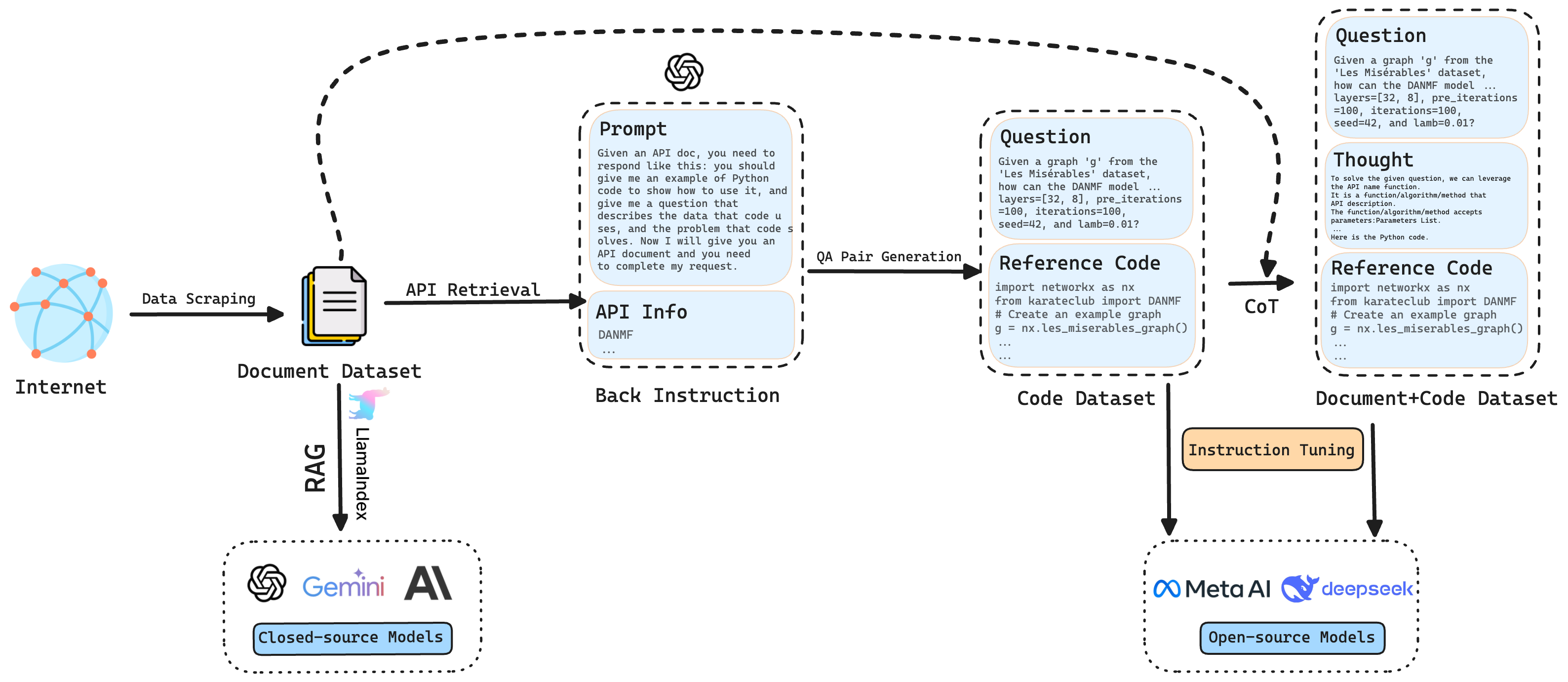}
    \caption{\textbf{The pipeline of \datasetname dataset construction and corresponding model enhancement.} We build the LLM4Graph dataset to improve the capabilities of LLMs in solving graph reasoning tasks through API calls. For the document dataset, we collect API documentation from the Internet. For code dataset, we automatically generate questions and corresponding codes with GPT-4 and API documentation. And for ``doc+code'' dataset, inspired by CoT reasoning, we add thought for each question, and combine the API documentation and the code dataset for construction.}
    \vspace{-1.35em}
    \label{fig:pipeline-dataset}
\end{figure}

To enhance the ability of LLMs to solve graph problems with Python APIs, we construct \datasetname datasets. Based on the \datasetname, we enhance both closed-source and open-source models.

\subsection{Datasets}

\textbf{Document dataset.} The document dataset is crawled from the official documents of the corresponding Python libraries. These documents can be directly used to enhance closed-source models via RAG. We also use these documents for generating code datasets. Specifically, each API corresponds to a specific entry of the document dataset, including the description, parameter list, return value list, code examples, and other contents. An example of API entry is shown in Appendix \ref{sec:doc-example}.

\iffalse
\begin{tcolorbox}[colback=gray!10, colframe=black, rounded corners, boxrule=1.5pt, fontupper=\normalsize, left=2mm, right=2mm, top=1mm, bottom=1mm]
    \textbf{Document QA pair Template1} \newline
    \textit{Question:} What's the \textit{field name} of \textit{API name} API in \textit{package name}?
    \newline
    \textit{Answer:}
    The document information about a particular API.\newline
    \textbf{Document QA pair Template2} \newline
    \textit{Question:}
    Can you tell me what API has \textit{field name}?
    \newline
    \textit{Answer:}The API is \textit{API name}.
\end{tcolorbox}

Here, the field name corresponds to a specific piece of information about the API, such as a Description or Example Code. The API name corresponds to the actual name of the API, and the package name corresponds to one of the six Python libraries.
\fi

\textbf{Code datasets.} We construct two code datasets in the form of QA pairs. The questions in both datasets are the same, but the answers differ. In the simpler dataset, each answer only contains Python code. Inspired by Chain of Thought (CoT)~\cite{Wei2022ChainOT}, each answer in the more complex dataset additionally includes relevant APIs and their documents as prefixes. This modification can facilitate open-source models to utilize document information more effectively. We name the above code datasets as Code (QA) and Doc+Code (QA), respectively. Unlike the hand-crafted benchmark, problems in the code datasets are automatically generated and each contains only one key API. 

Specifically, we first randomly select an API from the six Python libraries, and then employ GPT-4 turbo to generate example code for the API along with a corresponding question via back instruction~\cite{Wang2022SelfInstructAL}. In the Code (QA) dataset, each answer only contains the Python code in the generated json. In the Doc+Code (QA) dataset, we design a template to additionally incorporate the API document information into each answer. This allows a CoT process that first selects the possibly needed APIs, then provides the corresponding API information, and finally writes code to solve the problem. Besides, the questions in both code datasets need to undergo the role-play processing in Section~\ref{role-play-rephrasing} for diverse problem descriptions.  The prompt for back instruction and an example of the Doc+Code dataset can be found at Appendix \ref{sec:appendix-back-instruct-prompt} and \ref{sec:appendix-doc-code-example}:

\iffalse
\begin{tcolorbox}[colback=gray!10, colframe=black, rounded corners, boxrule=1.5pt, fontupper=\normalsize, left=2mm, right=2mm, top=1mm, bottom=1mm]
    \textbf{Question} \newline
    Given the karate club graph provided by NetworkX, how do we find its communities using the greedy modularity maximization method and then print out the sorted list of nodes for each community?
    \newline
    \textbf{Answer}
    \begin{verbatim}
import networkx as nx
# Create a graph
G = nx.karate_club_graph()
# Find communities in the graph
communities = nx.community.greedy_modularity_communities(G, 
weight='weight', resolution=1, cutoff=1, best_n=None)
# Print the sorted list of nodes in each community
for community in communities:
    print(sorted(community))
    \end{verbatim}
\end{tcolorbox}
\fi

\begin{table}[ht]
\vspace{-1.3em}
\centering
\caption{Statistics of \datasetname datasets.}
% \vspace{-0.15em}
\begin{tabular}{lccc}
\toprule
 & \textbf{Document}  & \textbf{Code (QA)} & \textbf{Doc+Code (QA)} \\ 
\midrule
Category 1 & 1,115  & 23,324 & 23,324 \\ 
Category 2 & 253   & 5,136  & 5,136  \\ 
Category 3 & 45      & 800   & 800   \\ 
\midrule
Total & 1,413 & 29,260 & 29,260 \\
\bottomrule
\end{tabular}
\end{table}
% \vspace{-1em}
% \end{wraptable}

\subsection{Models}
\label{sec:datasets-and-models-models}
We use the above datasets to improve various LLMs in handling graph analysis tasks. For closed-source models, we enhance them by retrieving relevant information from the document dataset as RAG. For open-source models, we fine-tune them using code datasets as instruction tuning.%, resulting in GraphAnalyst.

\textbf{Closed-source Models.} Due to the high difficulty of our \benchmarkname benchmark, mainstream LLMs (including Claude, GPT and Gemini) are not particularly strong in directly solving these problems. Therefore, before closed-source LLMs answer these questions, we retrieve the document information of top-$K$ relevant APIs based on Llamaindex~\cite{llamaindex}, allowing the models to learn through the context and enhance their performance. We explore $K=3,5,7$ to investigate the impact of RAG, and the models will be given the corresponding suffix as No RAG, RAG 3, RAG 5 or RAG 7.

\textbf{Open-source Models.} We use the two code datasets (\textit{i.e.,} Code and Doc+Code) to fine-tune Llama-3-8B-Instruct~\cite{llama3modelcard} and Deepseek-Coder-7B-instruct~\cite{deepseek-coder}, and consider the following model variants: (1) Code-only: LLMs are fine-tuned with the Code dataset. (2) Code + RAG 3/5/7: Code-only models are further equipped with RAG as closed-source ones. (3) Doc+Code: LLMs are fine-tuned with the Doc+Code dataset, and conduct inference by a corresponding two-step CoT reasoning process. In the first step, the model generates potential APIs based on the question. In the second step, we retrieve the API information and then provide the question along with the API information back to the model. The model then completes the remaining reasoning by writing code to solve the problem. The third group of models can maximize the use of document information to enhance the performance of open-source models, and significantly narrow the performance gap with closed-source large models.

\section{Experiment}

In this section, we present a comprehensive evaluation of our proposed benchmark, \benchmarkname, and the accompanying dataset, \datasetname. Our experiments assess the performance of both closed-source and open-source LLMs on graph analysis tasks. We demonstrate the limitations of current LLMs in handling structured graph data, and showcase the potential improvements achievable through the use of our datasets via RAG or instruction tuning. More results are shown in Appendix \ref{sec:experimental-results-of-all-models} and \ref{sec:experimental-results-of-best-performing-models}. 

\begin{table}[t]
\centering
\caption{Performance (\%) of different models on \benchmarkname.}
\resizebox{\textwidth}{!}{
\begin{tabular}{lcccccccc}
\toprule
 & \multicolumn{2}{c}{\textbf{Basic Graph Theory}} & \multicolumn{2}{c}{\textbf{Graph Statistical Learning}} & \multicolumn{2}{c}{\textbf{Graph Embedding}} & \multicolumn{2}{c}{\textbf{Overall}} \\
\cmidrule(lr){2-3} \cmidrule(lr){4-5} \cmidrule(lr){6-7} \cmidrule(lr){8-9}
\textbf{Model} & Pass Rate & Accuracy & Pass Rate & Accuracy & Pass Rate & Accuracy & Pass Rate & Accuracy \\ 
\midrule
Claude 3 Haiku  & 52.9 & 31.6 & 23.4 & 9.7  & 32.6 & 2.9  & 42.2 & 22.4 \\ % \midrule
Claude 3 Sonnet & 57.1 & 33.2 & 15.6 & 4.6  & 10.9 & 0.0  & 40.4 & 21.6 \\ % \midrule
Claude 3 Opus   & 69.2 & \underline{47.3} & 31.2 & \underline{15.1} & \textbf{47.8} & \textbf{14.5} & 55.9 & \underline{34.7} \\ \midrule
GPT-3.5         & 64.1 & 35.1 & 24.7 & 8.4  & 15.2 & 1.1  & 47.9 & 24.0 \\ % \midrule
GPT-4 turbo           & \textbf{72.4} & 42.1 & \underline{39.0} & 14.8 & \underline{41.3} & \underline{12.0} & \underline{59.6} & 31.2 \\ % \midrule
GPT-4o          & \underline{69.9} & \textbf{48.1} & \textbf{48.7} & \textbf{21.4} & 32.6 & 5.8  & \textbf{60.2} & \textbf{36.3} \\ \midrule
Gemini 1.0 Pro  & 48.7 & 27.7 & 9.1  & 1.7  & 19.6 & 3.3  & 34.2 & 17.7 \\ % \midrule
Gemini 1.5 Pro  & 59.6 & 37.2 & 21.4 & 6.6  & 13.0 & 1.8  & 43.9 & 24.8 \\ \midrule
Llama 3          & 36.5 & 17.3 & 12.3 & 3.8  & 15.2 & 0.4  & 27.3 & 11.7 \\ % \midrule
Deepseek Coder  & 56.1 & 33.8 & 30.5 & 9.8  & 30.4 & 7.6  & 46.1 & 24.2 \\
\bottomrule
\end{tabular}
}
\label{tab:main-result}
% \vspace{-1.3em}
\end{table}

\subsection{Experiment Settings and Baselines}

% \textbf{Settings}

In this work, we fine-tune two open-source models, Llama-3-8B-Instruct~\citep{llama3modelcard} and Deepseek-Coder-7B-Instruct-v1.5~\citep{DBLP:journals/corr/abs-2401-14196}, using the LLM4Graph dataset. For all models, we use alignment-handbook framework~\cite{Tunstall_The_Alignment_Handbook}, set the learning rate, epochs, and max length as 2e-4, 4, and 4096, running on NVIDIA 8*A100 GPUs (40GB). The training batch size is set to 1 and evaluation batch size is set to 4. For testing, we set temperature as 0 and maximum output tokens as 4096, ensuring the stable and reasonable generation.

As our baselines, we employ models without RAG, including the GPT series (GPT-3.5, GPT-4 turbo, GPT-4o)~\citep{DBLP:journals/corr/abs-2303-08774}, Claude 3 series (Haiku, Sonnet, Opus)~\citep{AnthropicClaude3}, and Gemini Pro (1.0 and 1.5)~\citep{DBLP:journals/corr/abs-2403-05530}, along with non-fine-tuned open-source smaller models Llama-3-8B-Instruct~\citep{llama3modelcard} and Deepseek-Coder-7B-Instruct-v1.5~\citep{DBLP:journals/corr/abs-2401-14196}. The detailed results are presented in Table ~\ref{tab:main-result}.

\subsection{Benchmarking LLMs on ProGraph}
We evaluate a variety of mainstream closed-source and open-source LLMs on the \benchmarkname, including GPT~\citep{DBLP:journals/corr/abs-2303-08774}, Claude~\citep{AnthropicClaude3}, Gemini~\citep{DBLP:journals/corr/abs-2403-05530}, Llama 3~\citep{llama3modelcard} and Deepseek Coder~\citep{DBLP:journals/corr/abs-2401-14196}.

The results, presented in Table \ref{tab:main-result}, indicate that none of the tested LLMs could effectively solve the problems in \benchmarkname. While GPT-4 turbo and its variant GPT-4o demonstrate relatively higher performance with an overall accuracy of 31.2\% and 36.3\% respectively, they still fall short in delivering satisfying accuracy across different categories. For instance, GPT-4 turbo achieves an accuracy of 42.1\% in Category 1 but only 14.8\% and 12.0\% in Categories 2 and 3, respectively. Similar trends are observed in other models. These results highlight the challenges faced by current LLMs in addressing structured graph data as human experts. This necessitates the development of specialized datasets and fine-tuning approaches to bridge this performance gap.

\subsection{Enhancing Closed-Source LLMs with Document Dataset}
To investigate the potential of enhancing LLMs' performance on graph analysis tasks, we utilize RAG techniques to extract relevant API information from the documents in \datasetname. This augmented context is then provided to the LLMs to assist in generating more accurate and effective solutions.

\begin{wrapfigure}{r}{0.5\textwidth}
    \vspace{-20pt}
    \centering
    \includegraphics[width=0.5\textwidth]{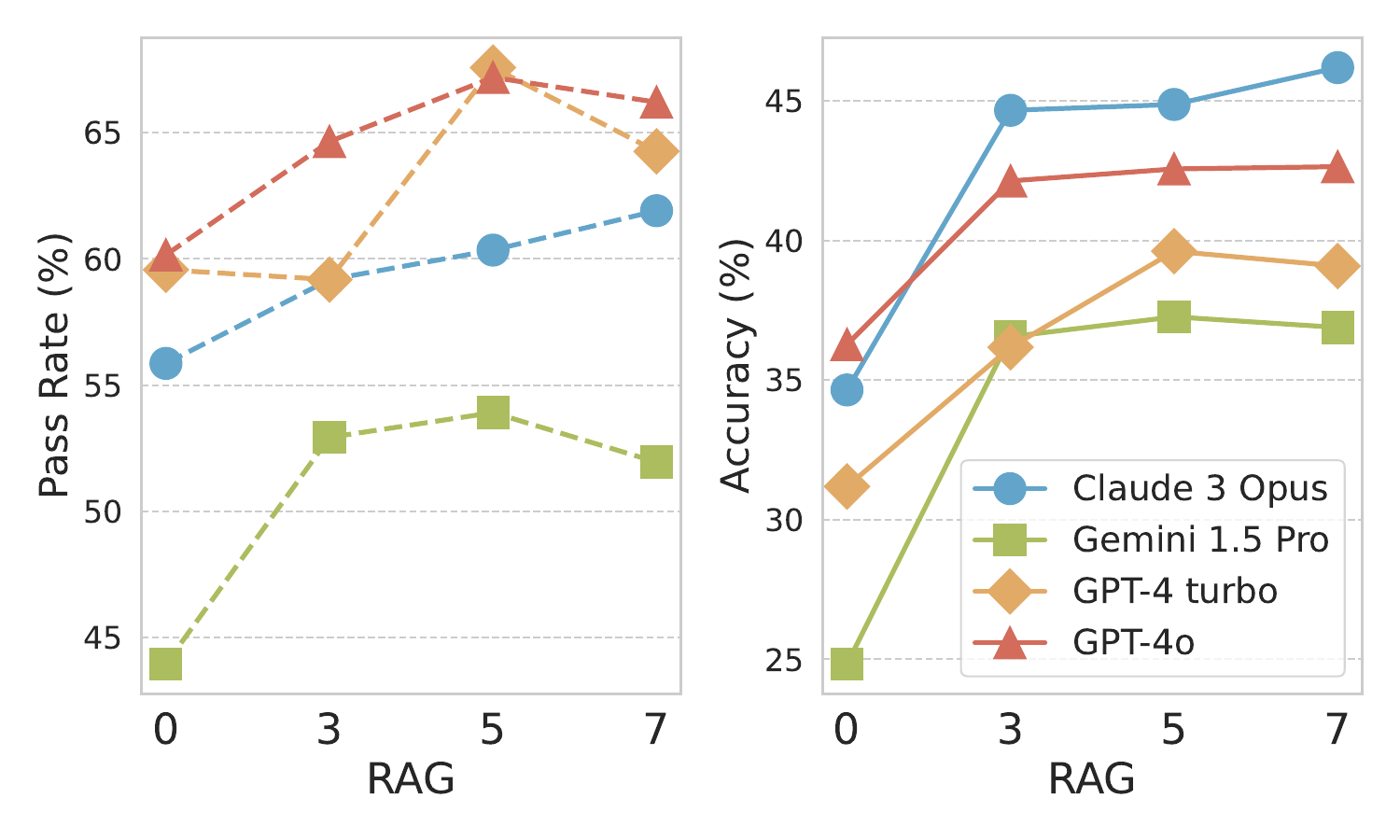}
    \caption{The pass rate (left) and accuracy (right) of closed-source models with RAG.}
    \label{fig:close-source-models-pass-rate-and-accuracy}
    \vspace{-20pt}
\end{wrapfigure}

Figure \ref{fig:close-source-models-pass-rate-and-accuracy} shows the performance gains for four top closed-source LLMs: GPT-4 turbo, GPT-4o, Claude 3 Opus, and Gemini 1.5 Pro. All models show significant boosts in pass rate and accuracy, with more than a 5\% improvement in accuracy. Gemini 1.5 Pro and Claude 3 Opus achieve over 10\% accuracy improvement with five retrieved APIs.
However, performance improvements plateau with additional API information, which may be attributed to the saturation of relevant information, where additional API details no longer contribute to further understanding and instead introduce redundancy or noise. This observation aligns with the findings of previous studies on RAG~\citep{DBLP:conf/nips/LewisPPPKGKLYR020}.

The effectiveness of the \datasetname in enhancing LLM capabilities on graph analysis tasks is evident from these results. By incorporating a RAG mechanism with the \datasetname, we demonstrate that it is possible to substantially improve the performance of closed-source LLMs without the need for extensive fine-tuning or architectural modifications. This approach offers a practical and efficient solution for adapting existing LLMs to handle structured graph data more effectively.

\subsection{Enhancing Open-Source LLMs with Code Dataset and Doc+Code Dataset}
\label{sec:experiments-open-source}

\begin{figure}[t]
    \centering
    \begin{subfigure}[b]{0.48\textwidth}
        \centering
        \includegraphics[width=\textwidth]{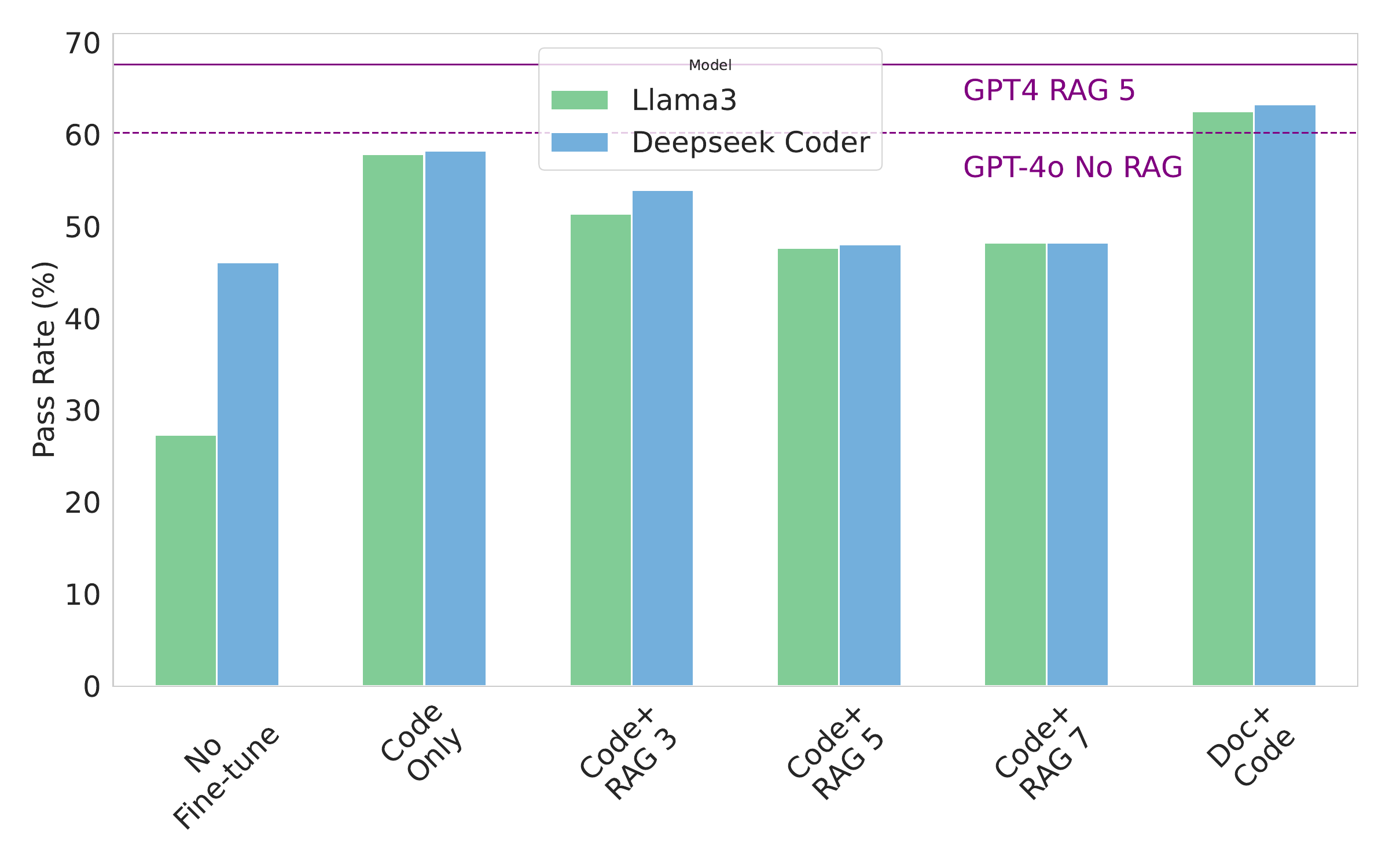}
        % \caption{Open Source Pass Rate}
        \label{fig:open-source-pass-rate}
    \end{subfigure}
    \hfill
    \begin{subfigure}[b]{0.48\textwidth}
        \centering
        \includegraphics[width=\textwidth]{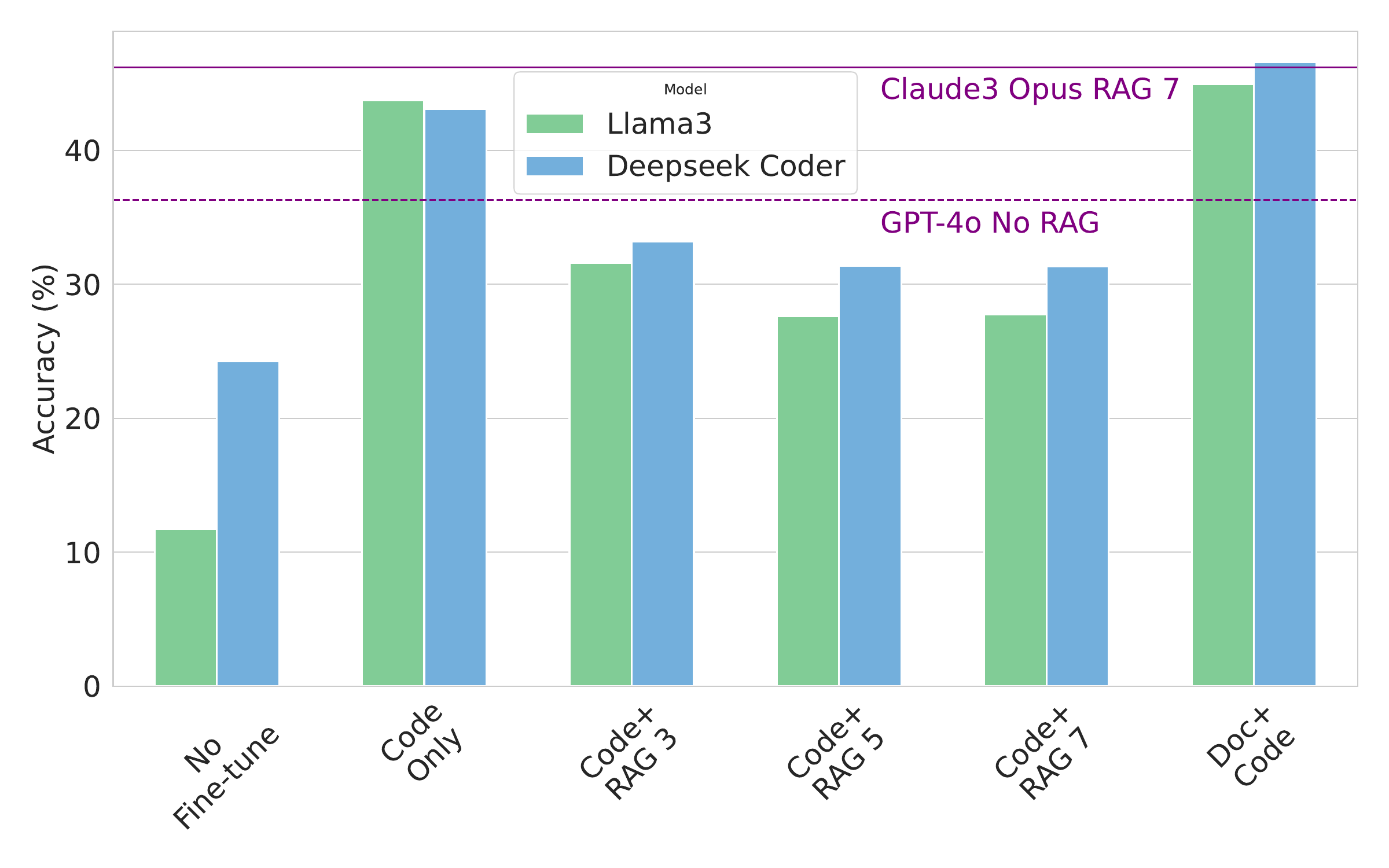}
        % \caption{Open Source Accuracy}
        \label{fig:open-source-accuracy}
    \end{subfigure}
    \vspace{-2em}
    \caption{The pass rate (left) and accuracy (right) of open-source models with instruction tuning.}
    \label{fig:open-source-models-pass-rate-and-accuracy}
\end{figure}

We investigate the potential of enhancing open-source LLMs' performance on graph analysis tasks by fine-tuning them using \datasetname and augmenting the models with RAG. Experiments are conducted on a general-purpose model, Llama-3-8B-Instruct~\citep{llama3modelcard}, and a model specifically designed for code generation, Deepseek-Coder-7B-Instruct-v1.5~\citep{DBLP:journals/corr/abs-2401-14196}. The results, presented in Figure \ref{fig:open-source-models-pass-rate-and-accuracy}, demonstrate that our \datasetname can significantly improve the performance of different types of open-source small models. The accuracy of both models, after being fine-tuned merely on the code data within \datasetname, substantially surpasses the best result from closed-source models without RAG. The Doc+Code setting further enhances the models' performance, leading to comparable or even better accuracy than the best result from closed-source model with RAG.

However, augmenting the open-source models fine-tuned on the code with RAG mechanism does not further improve the performance, and even leads to degraded performance. We hypothesize that this discrepancy may be attributed to the limited ability of these small models to process long text, hindering their utilization of the document information. The additional information provided by RAG may introduce confusion in understanding the problem statement and arriving at the correct solution. Overall, our proposed Doc+Code setting proves to be an effective means of integrating document information from \datasetname, significantly enhancing the accuracy of open-source models. We show that \datasetname can serve as effective data to enhance the model's code-generation capabilities and lead to better utilization of various graph libraries.

\section{Analysis}
\label{sec:analysis}
In this section, we present a comprehensive analysis of the performance of different models on the \benchmarkname. By grouping the benchmark based on categories, answer difficulties, and question type,  we aim to provide a granular exploration of the strengths and limitations of these models in handling graph analysis tasks. We also present the types of compilation errors made by different models. 

\subsection{Performance Analysis on Different Benchmark Groupings}
To gain a deeper understanding of the capabilities of LLMs and fine-tuned smaller models presented in Section \ref{sec:experiments-open-source}, we analyze their performances on \benchmarkname from three different perspectives: task category, answer difficulty and question type.

\begin{figure}[t]
    \centering
    \begin{subfigure}[b]{0.34\textwidth}
        \centering
        \includegraphics[width=\textwidth]{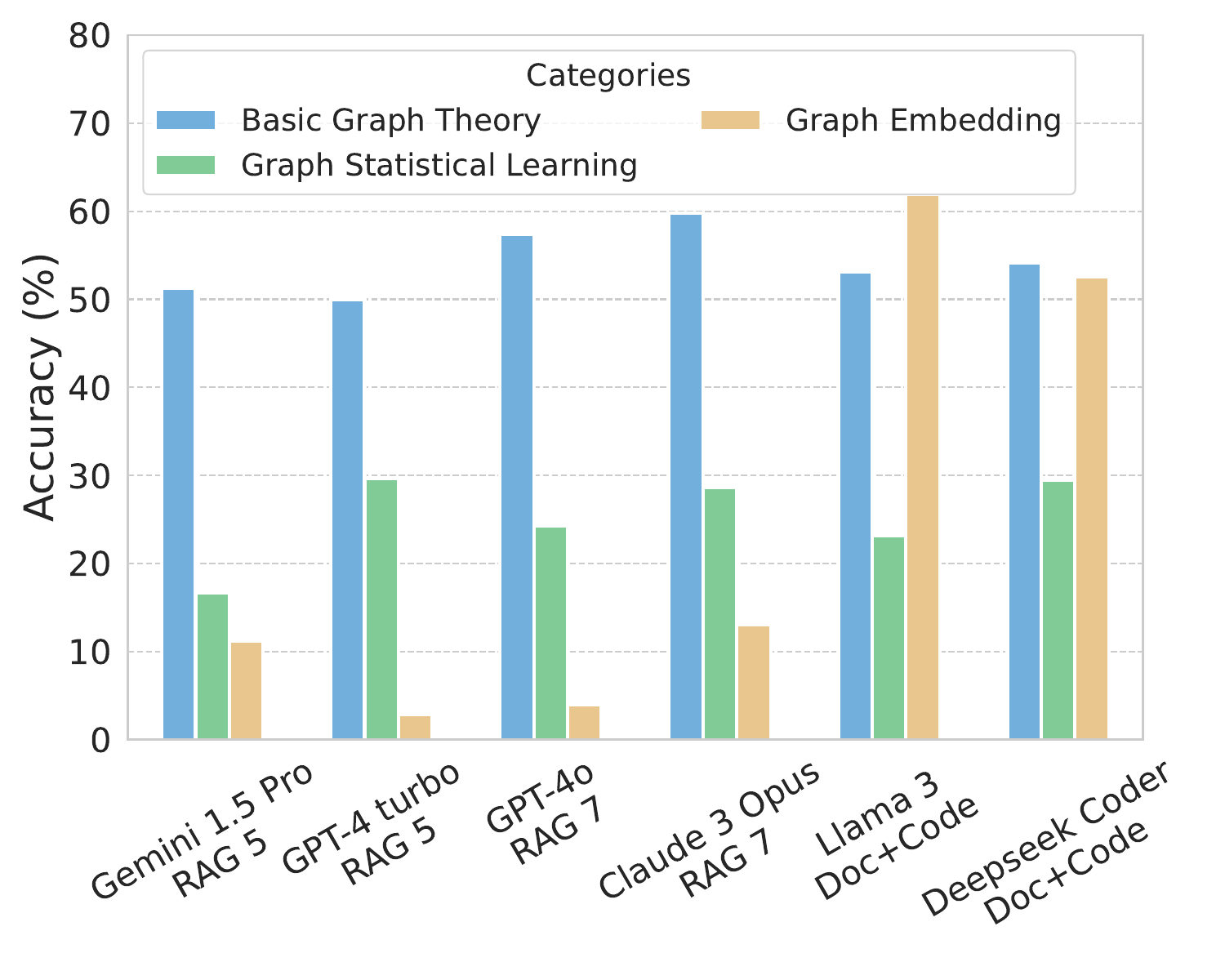}
        \vspace{-1.5em}
        \caption{Task Category}
        \label{fig:task-acc-category}
    \end{subfigure}
    \hspace{-1em}
    \begin{subfigure}[b]{0.34\textwidth}
        \centering
        \includegraphics[width=\textwidth]{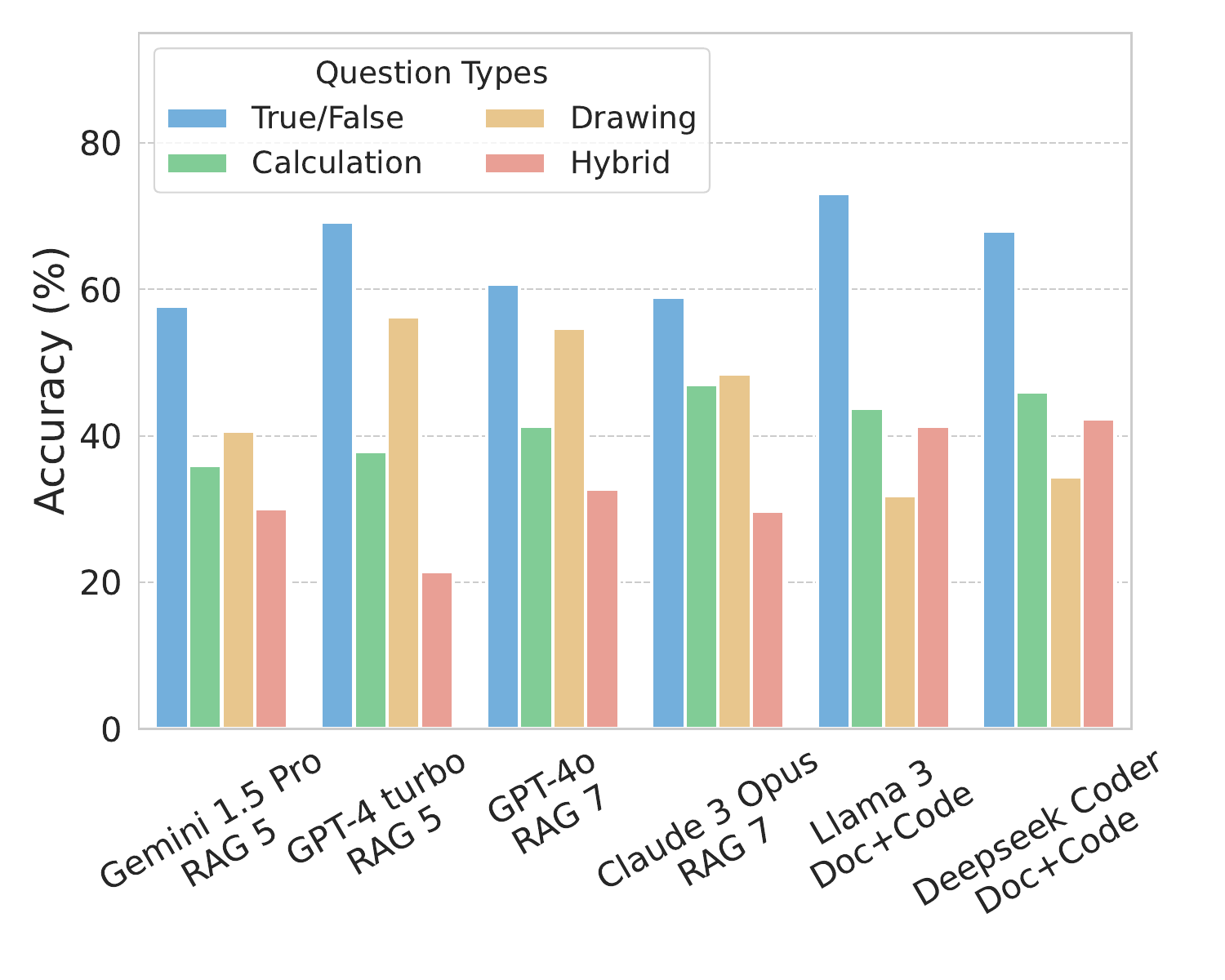}
        \vspace{-1.5em}
        \caption{answer difficulty}
        \label{fig:acc-problem-type}
    \end{subfigure}
    \hspace{-1em} 
    \begin{subfigure}[b]{0.34\textwidth}
        \centering
        \includegraphics[width=\textwidth]{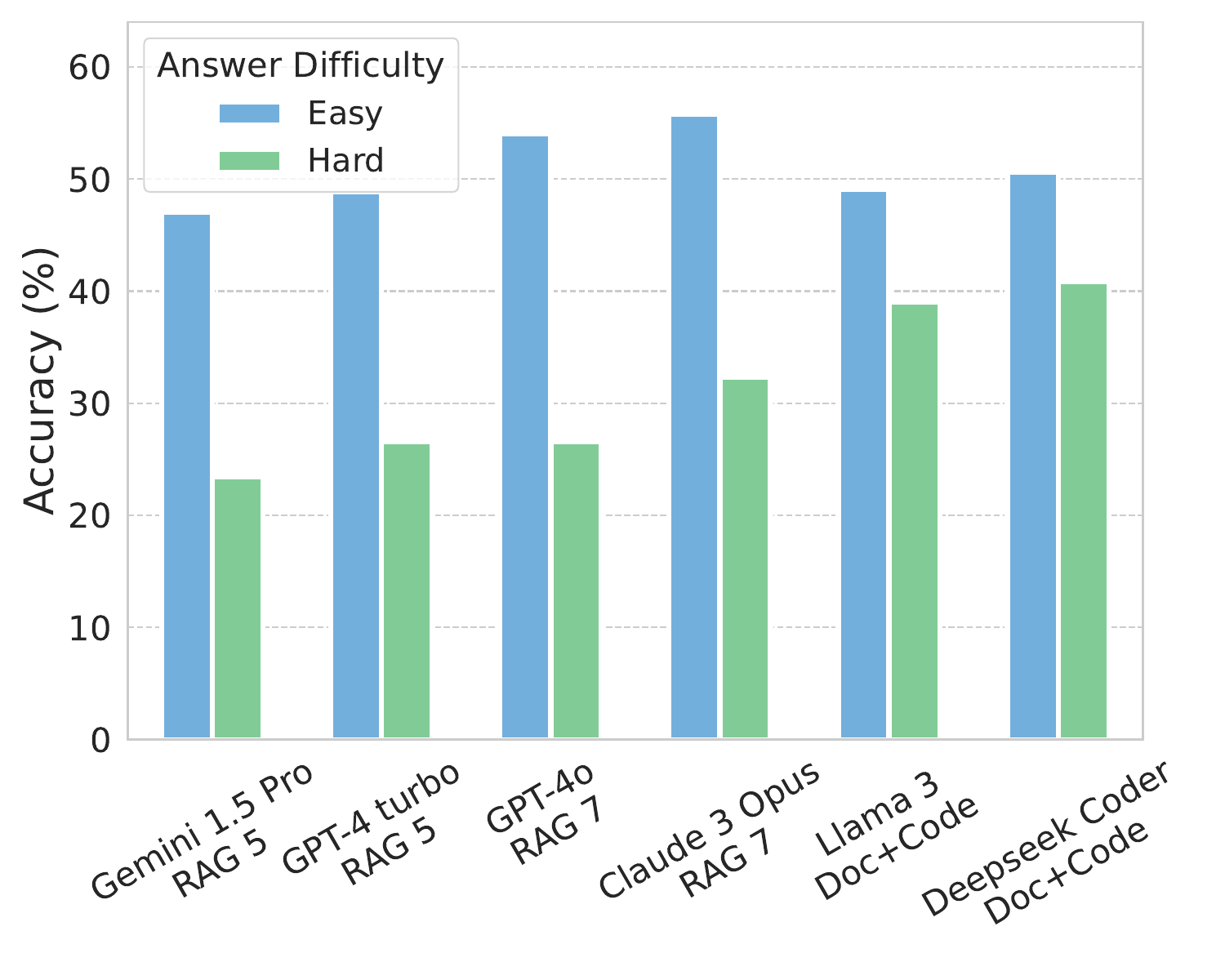}
        \vspace{-1.5em} 
        \caption{question type}
        \label{fig:acc-answer-difficulty}
    \end{subfigure}
    \caption{The performance of six best-performing models on different groupings of \benchmarkname.}
    \vspace{-1.5em}
    \label{fig:performance-different-groupings}
\end{figure}

\textbf{Task Category.}
We analyze the model performance based on different categories in the \benchmarkname, as shown in Figure \ref{fig:task-acc-category}. Mainstream LLMs and fine-tuned smaller models exhibit similar performance on graph theory and graph statistical learning tasks. However, a significant disparity is observed in their performance on graph embedding tasks, where fine-tuned smaller models substantially outperform RAG-enhanced large models. This observation suggests that not all graph analysis tasks can be easily handled by closed-source LLMs without further fine-tuning. More complex and challenging tasks still require fine-tuning for effective learning. 

\textbf{Answer Difficulty.}
We further examine the model performance based on different answer difficulties, \textit{i.e.,} true/false, calculation, drawing, and hybrid. In Figure \ref{fig:acc-problem-type}, we plot the performance of different models on these answer difficulties separately. Mainstream LLMs excel in true/false and drawing types but struggle with calculation and hybrid ones. Fine-tuned smaller models demonstrate improvements across various answer difficulties, especially on the complex hybrid type, indicating the effectiveness of our proposed enhancement strategies.

\textbf{Question Type.} 
Lastly, we divide the \benchmarkname into two levels of difficulty: easy (involving only one API) and hard (involving multiple APIs). As shown in Figure \ref{fig:acc-answer-difficulty}, mainstream closed-source large models perform well on easy-level problems, with accuracies generally approaching or exceeding 50\%. However, their performance significantly deteriorates on hard-level ones, with the highest accuracy reaching only 32.5\%. This observation suggests that mainstream LLMs have limitations when the number of required APIs increases. In contrast, our fine-tuned models demonstrate significantly higher accuracy on hard-level problems, approaching or exceeding 40\%, yielding approximately an improvement of 8\% compared to the best closed-source LLM. Note that \datasetname only contains data involving one API. Still, the models fine-tuned on \datasetname show strong generalizability on problems requiring multiple APIs.

\subsection{Compilation Error Analysis}
\begin{figure}[htbp]
\centering
\includegraphics[width=\textwidth]{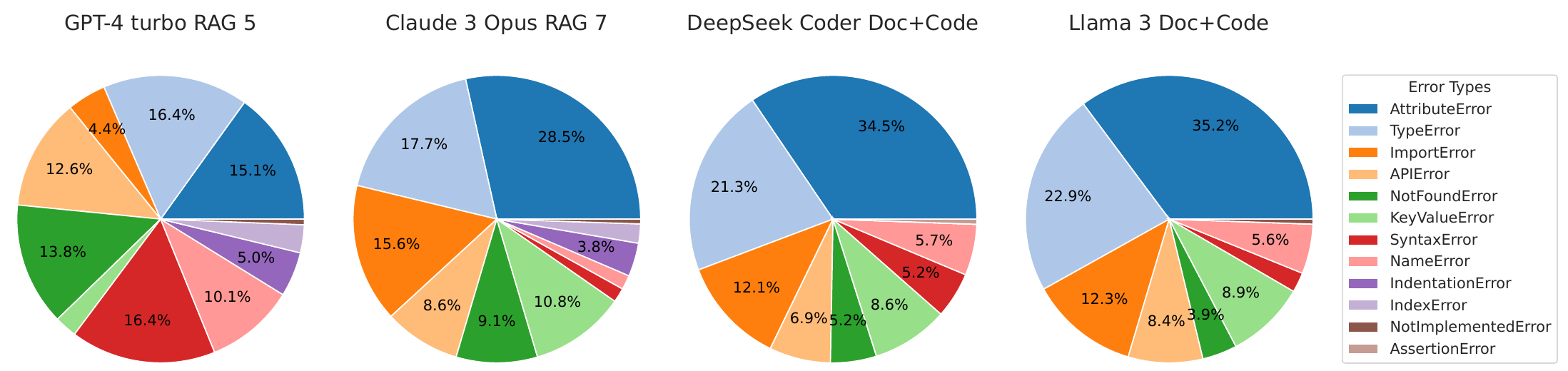}
\caption{Compilation error statistics for four best-performing models.}
\vspace{-2em}
\label{fig:error-analysis}
\end{figure}

To gain insights into the types of compilation errors made by different models, we conduct an error analysis on the best-performing closed-source models (GPT-4 turbo RAG 5 and Claude 3 Opus RAG 7) and fine-tuned open-source small models (DeepSeek Coder Doc+Code and Llama 3 Doc+Code). As shown in Figure \ref{fig:error-analysis}, we categorize the errors into ten distinct types to identify patterns and differences in the error distributions among these models.

Our analysis reveals that closed-source models exhibit a low similarity in their error cases, suggesting that they possess varying coding capabilities. For instance, GPT-4 turbo often makes SyntaxError, but rarely ImportError, which is contrary to Claude 3 Opus. The fine-tuned open-source small models exhibit a high similarity in their error distributions, with AttributeError being the most dominant. 

The error analysis also highlights some common challenges faced by all models, such as AttributeError and TypeError, suggesting that models may have difficulty in memorizing and understanding the attribute of class objects from various python libraries, and the type of returned results from different functions. Interestingly, the fine-tuned models have a notably lower percentage of SyntaxError compared to the closed-source models, indicating that further fine-tuning on the \datasetname helps the models learn better code syntax and structure. 

\section{Conclusion}

In this paper, we introduce \benchmarkname, a novel and challenging benchmark for evaluating LLMs in graph analysis using external APIs. Current LLMs achieve only 36\% accuracy, revealing their limitations. To bridge this gap, we further construct \datasetname, a dataset with crawled documents and auto-generated codes based on popular graph libraries. The datasets can help improve the accuracy of both closed-source and open-source LLMs by 11-32\% through RAG and instruction tuning. Our work highlights the potential of enhancing LLMs with our \datasetname, offering valuable resources for advancing LLM capabilities in structured data analysis. Discussions about limitations and boarder impacts can be found in Appendix~\ref{limitation}.

\section{Acknowledgements}

This work is supported by the National Natural Science Foundation of China (No.62192784, 62236004), the National Key R\&D Program of China (No.2022ZD0116312), Young Elite Scientists Sponsorship Program (No.2023QNRC001) by CAST, and Tsinghua University Initiative Scientific Research Program.

\medskip
\bibliographystyle{plainnat}
\bibliography{reference}

\newpage
\appendix

% \section{Appendix}

\section{Limitation and Boarder Impact}
\label{limitation}
\textbf{Limitation.} Due to the version iterations of GPT, the results based on automated evaluation might show slight differences compared to those presented in this paper. The proposed \benchmarkname benchmark contains 512 hand-crafted problems, which can only reflect a core subset of graph analysis tasks. Note that our dataset includes nearly 30,000 auto-generated instances. Thus, in future work, we can expand the benchmark by manually selecting high-quality cases from the dataset. Admittedly, the benchmark still presents certain differences from real-world scenarios, \textit{e.g.,} we did not consider a multi-turn task-solving process.

\textbf{Boarder Impact.} The inherent complexity of graph data in real-world scenarios demands professional coding capabilities, making it a compelling testbed for evaluating the structured data handling abilities of LLMs. We believe this work can shed light on further researches toward more powerful general intelligence with structured data understanding.

% here

\newpage

\section{Role-play Rephrasing Example}
\label{sec:role-play-rephrasing-example}

\begin{tcolorbox}[colback=gray!10, colframe=black, rounded corners, boxrule=1.5pt, fontupper=\normalsize, left=2mm, right=2mm, top=1mm, bottom=1mm]
    \textbf{Original Question} \newline
    I have a graph with an edge set [(1, 2), (1, 3), (2, 3), (2, 4), (3, 5), (4, 5)], can you help me compute node connectivity for all pairs of nodes and print the node connectivity for each pair?
    \newline
    \textbf{Role-Play Rephrasing Prompt}  \newline
The above question is too abstract. 
Please role-play based on the following professional information, 
mimic their everyday language and tone,
and give a background from real life (more than 4 sentences).
Restate the problem, ensuring that the semantics remain unchanged.
Notes: If the original question has a file, you must give the file name in your new question.
    \newline
    \textbf{Professional Information} \newline
    \textit{Ecologist:} Studies the relationships between organisms and their environments, including ecosystems, habitats, and biodiversity.
    \newline
    \textbf{Rephrased Question} \newline
Imagine we're examining a simplified model of an ecosystem where the nodes represent different species, and the edges indicate direct ecological interactions between them, such as predator-prey or symbiotic relationships. For our current study, we've mapped out a series of interactions as follows: [(1, 2), (1, 3), (2, 3), (2, 4), (3, 5), (4, 5)]. 

To fully understand the resilience of our ecosystem, we need to determine the species connectivity, which reflects how each pair of species is interconnected through these interactions. Can we analyze our network to reveal the minimum number of species that would need to be removed to disrupt the direct connection between any two species in this web? I'd like to have a measurable indication of node connectivity for each unique pair of species within our ecosystem based on the interactions detailed above. This will provide us with valuable insights into the robustness of their relationships and, by extension, the stability of our ecosystem.
\end{tcolorbox}

\section{Document Dataset Example}
\label{sec:doc-example}
\begin{tcolorbox}[colback=gray!10, colframe=black, rounded corners, boxrule=1.5pt, fontupper=\normalsize, left=2mm, right=2mm, top=1mm, bottom=1mm]
    % \caption{test}
    \textbf{Section ID} triadic\_census
    \newline
    \textbf{Description} \newline
    Determines the triadic census of a directed graph. The triadic census is a count of how many ...
    \newline
    \textbf{Field List} \newline
    \textit{Parameters:} G : digraph; nodelist : list
    \newline
    \textit{Returns:} census : dict
    \newline
    \textit{Raises:} ValueError
    \newline
    \textbf{Rubrics} \newline
    \textit{Notes:} This algorithm has complexity O(m) where m is the number of edges in the graph ...
    \newline
    \textit{References:}  [1]Vladimir Batagelj and Andrej Mrvar, A subquadratic triad census algorithm ...
    \textit{Examples:}
    \vspace{-0.5em}
    \begin{verbatim}
        G=nx.DiGraph([(1,2),(2,3),(3,1),(3,4),(4,1),(4,2)])
        ...
    \end{verbatim}
    \vspace{-1.5em}
\end{tcolorbox}

\section{Prompt for GPT Back-Instruction}
\label{sec:appendix-back-instruct-prompt}
\begin{tcolorbox}[colback=gray!10, colframe=black, rounded corners, boxrule=1.5pt, fontupper=\normalsize, left=2mm, right=2mm, top=1mm, bottom=1mm]
    %\textbf{Few Shot} \newline
    %\textit{Standard Code QA 1}
    %\newline
    %\textit{Standard Code QA 2}
    %\newline
    %\textit{Standard Code QA 2}
    %\newline
    \textbf{Back Instruction Prompt} \newline
Given an API doc, you need to respond like this: you should give me an example of Python code to show how to use it, and give me a question that describes the data that code uses, and the problem that code solves. Now I will give you an API document and you need to complete my request.
    \newline
    %\textbf{API Document} \newline
    ** A complete API documentation entry **
    \newline
    %\textbf{JSON Output} \newline
Generate a JSON object like this: \{API: , example Python code: ,   question: , \}
\end{tcolorbox}

\newpage
\section{ProGraph Benchmark Example}
\label{sec:prograph-benchmark-example}

\begin{table}[ht]
\centering
\caption{Benchmark Example}
\begin{tabular}{p{3cm} p{10cm}}
\toprule
\textbf{Conponent} & \textbf{Example} \\ 
\midrule
Annotated Question & Given a graph with edge set [(1, 2), (1, 3), (2, 4), (3, 4), (4, 5), (5, 6), (5, 7), (6, 7)], can you Color a graph using largest\_first coloring strategy of greedy graph coloring?

Notes: You need to print the result like this.
\begin{verbatim}
for node, color in coloring.items():
    print(f"Node {node}: Color {color}")
\end{verbatim}
\\
Rephrased Question & As a Project Coordinator, you're tasked with the organization of a network of collaborative tasks, each represented by connections between different points, or 'nodes,' in a particular sequence. This sequence is comprised of pairs indicating task interdependencies: [(1, 2), (1, 3), (2, 4), (3, 4), (4, 5), (5, 6), (5, 7), (6, 7)]. The challenge lies in assigning 'colors,' or categories, to these tasks in a manner that adjacent tasks (those directly connected) are not categorized the same this ensures a clear division of responsibilities and resources.

Your goal is to utilize the "largest\_first" strategy within the greedy coloring algorithm for optimal organization. This strategic approach selects nodes based on their descending degree, that is, beginning with the node with the most connections. By categorizing in this method, you're likely to minimize the total number of categories used.

Upon completion of this process, we need to visualize our task categorization. Could you please format the output as a sequence of print statements in Python, displaying the nodes alongside their assigned categories? The output should follow this structure for clarity:
\begin{verbatim}
for node, color in coloring.items():
    print(f"Node {node}: Color {color}")
\end{verbatim}
This simplification will effectively translate the abstract graph coloring problem into a practical scenario suitable for organizing our network of tasks. \\
Answer & Node 4: Color 0
Node 5: Color 1
Node 1: Color 0
Node 2: Color 1
Node 3: Color 1
Node 6: Color 0
Node 7: Color 2
 \\
Reference Code & \begin{verbatim}
import networkx as nx
from networkx.algorithms.coloring import greedy_color
# Create a sample graph
G = nx.Graph()
edges = [(1, 2), (1, 3), (2, 4), (3, 4),
(4, 5), (5, 6), (5, 7), (6, 7)]
G.add_edges_from(edges)
# Apply the largest first coloring strategy
coloring = greedy_color(G, strategy="largest_first")
for node, color in coloring.items():
    print(f"Node {node}: Color {color}")    
\end{verbatim}
 \\
Key APIs & greedy\_color \\
\bottomrule
\end{tabular}
\end{table}
\newpage

\newpage

\section{Doc+Code Dataset Example}
\label{sec:appendix-doc-code-example}

% complex code dataset
\begin{tcolorbox}[colback=gray!10, colframe=black, rounded corners, boxrule=1.5pt, fontupper=\normalsize, left=2mm, right=2mm, top=1mm, bottom=1mm]
    \textbf{Question} \newline
    Given the karate club graph provided by NetworkX, how do we find its communities using the greedy modularity maximization method and then print out the sorted list of nodes for each community?
    \newline
    \textbf{Answer} \newline
To solve the given question, 
we can leverage {greedy\_modularity\_communities} function.  \newline
It is a function that {"Find communities in G using greedy modularity maximization.
This function uses Clauset-Newman-Moore greedy modularity maximization [2] to find the community partition with the largest modularity.
Greedy modularity maximization begins with each node in its own community and repeatedly joins the pair of communities that lead to the largest modularity until no further increase in modularity is possible (a maximum). Two keyword arguments adjust the stopping condition. cutoff is a lower limit on the number of communities so you can stop the process before reaching a maximum (used to save computation time). best\_n is an upper limit on the number of communities so you can make the process continue until at most n communities remain even if the maximum modularity occurs for more. To obtain exactly n communities, set both cutoff and best\_n to n.
This function maximizes the generalized modularity, where resolution is the resolution parameter, often expressed as (gamma). See modularity() ."}. \newline\newline
The function accepts parameters: \newline
"Parameters:": \{ \newline
\hspace*{4em} "G : NetworkX graph": "", \newline
\hspace*{4em} "weight : string or None, optional (default=None)": "The name of an edge attribute that holds the numerical value used as a weight.  If None, then each edge has weight 1. The degree is the sum of the edge weights adjacent to the node.", \newline
\hspace*{4em} "resolution : float, optional (default=1)": "If resolution is less than 1, modularity favors larger communities. Greater than 1 favors smaller communities.", \newline
\hspace*{4em} "cutoff : int, optional (default=1)": "A minimum number of communities below which the merging process stops. The process stops at this number of communities even if modularity is not maximized. The goal is to let the user stop the process early. The process stops before the cutoff if it finds a maximum of modularity.", \newline
\hspace*{4em} "best\_n : int or None, optional (default=None)": "A maximum number of communities above which the merging process will not stop. This forces community merging to continue after modularity starts to decrease untilbest\_ncommunities remain. If None, force it to continue beyond a maximum."\} \newline
\newline
The function returns:
"Returns:": \{ \newline
\hspace*{4em} "communities: list": "A list of frozensets of nodes, one for each community. Sorted by length with largest communities first."
            \} \newline
The function's path is {"networkx.community.greedy\_modularity\_communities"}. \newline

By using the info above, we can write a Python code to solve this problem.

Here is the Python code.
    \begin{verbatim}
import networkx as nx
# Create a graph
G = nx.karate_club_graph()
# Find communities in the graph
communities = nx.community.greedy_modularity_communities(G, 
weight='weight', resolution=1, cutoff=1, best_n=None)
# Print the sorted list of nodes in each community
for community in communities:
    print(sorted(community))
    \end{verbatim}
\end{tcolorbox}

\clearpage

\section{Experimental Results of All Models}
\label{sec:experimental-results-of-all-models}

\begin{table}[ht]
\centering
\caption{Performance (\%) of closed-source models regarding different task categories.}
\scalebox{0.68}{
\begin{tabular}{cc|cc|cc|cc|cc}
\toprule
\multicolumn{2}{c|}{\multirow{2}{*}{Model}} & \multicolumn{2}{c|}{Graph Theory} & \multicolumn{2}{c|}{Graph Statistical Learning} & \multicolumn{2}{c|}{Graph Embedding} & \multicolumn{2}{c}{Overall} \\ \cmidrule(lr){3-4} \cmidrule(lr){5-6} \cmidrule(lr){7-8} \cmidrule(lr){9-10}
\multicolumn{2}{c|}{} & Pass Rate & Accuracy & Pass Rate & Accuracy & Pass Rate & Accuracy & Pass Rate & Accuracy \\ 
\midrule
\multirow{4}{*}{Claude 3 Haiku}
 & No RAG & 52.9 & 31.6 & 23.4 & 9.7  & 32.6 & 2.9  & 42.2 & 22.4 \\
 & RAG 3  & 68.9 & 47.7 & 22.1 & 11.4 & 23.9 & 1.1  & 50.8 & 32.6 \\
 & RAG 5  & 63.5 & 44.4 & 29.9 & 16.4 & 15.2 & 2.5  & 49.0 & 32.2 \\
 & RAG 7  & 65.4 & 51.0 & 25.3 & 15.2 & 17.4 & 6.5  & 49.0 & 36.2 \\ \midrule
 \multirow{4}{*}{Claude 3 Sonnet}
 & No RAG & 57.1 & 33.2 & 15.6 & 4.6 & 10.9 & 0.0 & 40.4 & 21.6 \\
 & RAG 3  & 63.5 & 45.8 & 13.6 & 7.6 & 19.6 & 5.8 & 44.5 & 30.7 \\
 & RAG 5  & 63.5 & 45.5 & 16.2 & 9.7 & 21.7 & 4.7 & 45.9 & 31.1 \\
 & RAG 7  & 66.4 & 50.0 & 25.3 & 12.3 & 21.7 & 4.3 & 50.0 & 34.6 \\ \midrule
\multirow{4}{*}{Claude 3 Opus} 
 & No RAG & 69.2 & 47.3 & 31.2 & 15.1 & 47.8 & \underline{14.5} & 55.7 & 34.7 \\
 & RAG 3  & 74.4 & \underline{59.4} & 39.6 & 28.3 & 21.7 & 0.0  & 59.2 & 44.7 \\
 & RAG 5  & 73.4 & 56.4 & 39.6 & \underline{28.8} & 41.3 & \textbf{20.7} & 60.4 & \underline{44.9} \\
 & RAG 7  & \underline{75.6} & \textbf{59.8} & 42.9 & 28.6 & 32.6 & 13.0 & 61.9 & \textbf{46.2} \\ \midrule
\multirow{4}{*}{GPT-3.5}        
 & No RAG & 64.1 & 35.1 & 24.7 & 8.4  & 15.2 & 1.1 & 47.9 & 24.0 \\
 & RAG 3  & 67.0 & 44.3 & 32.5 & 12.0 & 41.3 & 5.1 & 54.3 & 31.1 \\
 & RAG 5  & 64.4 & 45.2 & 33.1 & 16.2 & 43.5 & 5.4 & 53.1 & 32.9 \\
 & RAG 7  & 64.7 & 45.8 & 33.8 & 15.9 & 37.0 & 3.3 & 52.9 & 33.0 \\ \midrule
\multirow{4}{*}{GPT-4 turbo}          
 & No RAG & 72.4 & 42.1 & 39.0 & 14.8 & 41.3 & 12.0 & 59.6 & 31.2 \\
 & RAG 3  & 74.7 & 48.5 & 40.3 & 21.4 & 17.4 & 2.2  & 59.2 & 36.2 \\
 & RAG 5  & \underline{75.6} & 50.0 & \textbf{56.5} & \textbf{29.7} & 50.0 & 2.9  & \textbf{67.6} & 39.6 \\
 & RAG 7  & 74.7 & 51.3 & 46.8 & 23.8 & 52.2 & 7.6  & 64.3 & 39.1 \\ \midrule
\multirow{4}{*}{GPT-4o}         
 & No RAG & 69.9 & 48.1 & 48.7 & 21.4 & 32.6 & 5.8  & 60.2 & 36.3 \\
 & RAG 3  & 73.7 & 55.5 & 51.3 & 24.7 & 47.8 & 9.8 & 64.7 & 42.1 \\
 & RAG 5  & 74.7 & 54.8 & \underline{55.2} & 27.8 & \textbf{56.5} & 9.1 & \underline{67.2} & 42.6 \\
 & RAG 7  & \textbf{76.9} & 57.4 & 48.1 & 24.3 & \underline{54.4} & 4.0 & 66.2 & 42.7 \\ \midrule
\multirow{4}{*}{Gemini 1.0 Pro} 
 & No RAG & 48.7 & 27.7 & 9.1  & 1.7 & 19.6 & 3.3 & 34.2 & 17.7 \\
 & RAG 3  & 61.5 & 47.4 & 16.2 & 7.5 & 15.2 & 2.2 & 43.8 & 31.3 \\
 & RAG 5  & 62.2 & 44.4 & 15.6 & 6.8 & 13.0 & 0.0 & 43.8 & 29.1 \\
 & RAG 7  & 64.4 & 45.3 & 15.6 & 5.8 & 19.6 & 0.0 & 45.7 & 29.4 \\ \midrule
\multirow{4}{*}{Gemini 1.5 Pro} 
 & No RAG & 59.6 & 37.2 & 21.4 & 6.6  & 13.0 & 1.8  & 44.0 & 24.8 \\
 & RAG 3  & 70.2 & 52.0 & 24.7 & 12.0 & 30.4 & 13.8 & 52.9 & 36.6 \\
 & RAG 5  & 71.2 & 51.3 & 29.2 & 16.7 & 19.6 & 11.2 & 53.9 & 37.3 \\
 & RAG 7  & 70.5 & 51.9 & 23.4 & 15.3 & 21.7 & 7.3  & 52.0 & 36.9 \\ 
\bottomrule
\end{tabular}
}
\end{table}

\begin{table}[ht]
\centering
\caption{Performance (\%) of open-source models regarding different task categories.}
\scalebox{0.6}{
\begin{tabular}{cc|cc|cc|cc|cc}
\toprule
\multicolumn{2}{c|}{\multirow{2}{*}{Model}} & \multicolumn{2}{c|}{Graph Theory} & \multicolumn{2}{c|}{Graph Statistical Learning} & \multicolumn{2}{c|}{Graph Embedding} & \multicolumn{2}{c}{Overall} \\ \cmidrule(lr){3-4} \cmidrule(lr){5-6} \cmidrule(lr){7-8} \cmidrule(lr){9-10}
\multicolumn{2}{c|}{} & Pass Rate & Accuracy & Pass Rate & Accuracy & Pass Rate & Accuracy & Pass Rate & Accuracy \\ 
\midrule
\multirow{9}{*}{Llama 3}         
 & No Fine-tune        & 36.5 & 17.3 & 12.3 & 3.8  & 15.2 & 0.4  & 27.3 & 11.7 \\

 & Code Only          & 61.2 & 46.7 & \textbf{49.4} & \textbf{36.6} & 63.0 & 47.5 & 57.8 & 43.8 \\

 & Code+RAG 3       & 51.6 & 30.1 & \underline{47.4} & 30.9 & 63.0 & 44.2 & 51.4 & 31.6 \\
 & Code+RAG 5       & 47.8 & 25.2 & 44.8 & 28.9 & 56.5 & 40.2 & 47.7 & 27.6 \\
 & Code+RAG 7       & 47.1 & 25.1 & 46.8 & 30.6 & 60.9 & 36.2 & 48.2 & 27.7 \\
 & Doc+Code & \underline{69.6} & \underline{53.2} & 42.9 & 23.2 & \textbf{80.4} & \textbf{62.0} & \underline{62.5} & \underline{44.9} \\ \cmidrule(lr){1-10}
\multirow{9}{*}{Deepseek Coder} 
 & No Fine-tune        & 56.1 & 33.8 & 30.5 & 9.9  & 30.4 & 7.6  & 46.1 & 24.2 \\
 & Code Only          & 62.5 & 46.9 & \underline{47.4} & \underline{33.4} & 65.2 & 49.6 & 58.2 & 43.1 \\
 & Code+RAG 3       & 59.0 & 35.9 & 46.8 & 28.8 & 43.5 & 29.4 & 53.9 & 33.2 \\
 & Code+RAG 5       & 52.9 & 32.0 & 43.5 & 32.3 & 30.4 & 24.3 & 48.1 & 31.4 \\
 & Code+RAG 7       & 51.6 & 33.6 & \textbf{49.4} & 31.4 & 21.7 & 15.9 & 48.2 & 31.4 \\
 & Doc+Code  & \textbf{71.2} & \textbf{54.1} & 46.1 & 29.8 & \underline{67.4} & 52.5 & \textbf{63.3} & \textbf{46.6} \\ 
\bottomrule
\end{tabular}
}
\end{table}

\begin{table}[ht]
\centering
\caption{Performance (\%) of closed-source models regarding different answer difficulties.}
\scalebox{0.68}{
\begin{tabular}{cc|cc|cc|cc|cc}
\toprule
\multicolumn{2}{c|}{\multirow{2}{*}{Model}} & \multicolumn{2}{c|}{True/False} & \multicolumn{2}{c|}{Drawing} & \multicolumn{2}{c|}{Calculation} & \multicolumn{2}{c}{Hybrid} \\ \cmidrule(lr){3-4} \cmidrule(lr){5-6} \cmidrule(lr){7-8} \cmidrule(lr){9-10}
\multicolumn{2}{c|}{} & Pass Rate & Accuracy & Pass Rate & Accuracy & Pass Rate & Accuracy & Pass Rate & Accuracy \\ 
\midrule
\multirow{4}{*}{Claude 3 Haiku}
 & No RAG & 71.8 & 53.4 & 43.4 & 20.2 & 34.4 & 27.1 & 41.1 & 13.5  \\
 & RAG 3 & 71.8 & 52.6 & 52.5 & 30.9 & 53.1 & 46.9 & 41.1 & 22.3 \\
 & RAG 5  & 66.7 & 48.7 & 51.2 & 31.4 & 43.8 & 37.5 & 37.5 & 23.4 \\
 & RAG 7  & 64.1 & 56.4 & 49.4 & 34.9 & 44.6 & 24.1 & 53.1 & \textbf{48.4} \\ \midrule
\multirow{4}{*}{Claude 3 Sonnet}
 & No RAG & 56.4 & 38.5 & 43.9 & 21.6 & 21.9 & 15.1  & 28.6 & 13.7 \\
 & RAG 3 & \textbf{79.5} & 68.4 & 45.7 & 28.2 & 34.4 & 31.3  & 30.4 & 21.4 \\
 & RAG 5 & 66.7 & 53.8 & 46.0 & 29.6 & 40.63 & 35.4  & 39.3 & 22.4 \\
 & RAG 7 & 71.8 & 51.3 & 52.2 & 33.8 & 41.1 & 23.2  & 43.8 & \underline{43.8} \\ \midrule
\multirow{4}{*}{Claude 3 Opus}
 & No RAG & \textbf{79.5} & 66.7 & 57.1 & 33.0 & 37.5 & 28.1 & 50.0 & 27.5 \\
 & RAG 3  & \textbf{79.5} & \textbf{71.0} & 61.6 & 44.9 & 43.8 & 40.1 & 48.2 & 27.5 \\
 & RAG 5  & \underline{76.9} & 66.7 & 62.9 & \underline{45.0} & 40.6 & 37.0 & 55.4 & 33.4 \\
 & RAG 7  & 74.4 & 59.0 & 65.2 & \textbf{46.9} & 50.0 & 48.4 & 48.2 & 29.7 \\ \midrule
\multirow{4}{*}{GPT-3.5}
 & No RAG & 66.7 & 38.5 & 51.4 & 23.4 & 43.8 & 30.2 & 32.1 & 14.4 \\
 & RAG 3  & 74.4 & 56.4 & 55.6 & 29.8 & 43.8 & 35.4 & 51.8 & 19.8 \\
 & RAG 5  & 69.2 & 61.5 & 55.6 & 31.5 & 40.6 & 31.3 & 44.6 & 23.2 \\
 & RAG 7  & 56.4 & 41.0 & 56.1 & 32.5 & 50.0 & 47.9 & 51.8 & 21.9 \\ \midrule
\multirow{4}{*}{GPT-4 turbo}
 & No RAG & \underline{76.9} & 51.3 & 62.3 & 31.3 & 43.8 & 29.7 & 44.6 & 17.0 \\
 & RAG 3  & \underline{76.9} & 55.1 & 60.3 & 33.7 & \underline{62.5} & 52.1 & 48.2 & 31.1 \\
 & RAG 5  & \textbf{79.5} & \underline{69.2} & \textbf{69.4} & 37.9 & \underline{62.5} & \textbf{56.3} & 60.7 & 21.4 \\
 & RAG 7  & 71.8 & 59.0 & 66.5 & 38.3 & \textbf{65.6} & 53.1 & 57.1 & 22.4 \\ \midrule
\multirow{4}{*}{GPT-4o}
 & No RAG & 71.8 & 53.9 & 61.3 & 34.7 & 56.3 & 44.3 & \underline{62.5} & 30.7 \\
 & RAG 3  & \textbf{79.5} & 65.8 & 65.2 & 39.4 & 56.3 & \underline{54.7} & \textbf{69.6} & 37.4 \\
 & RAG 5  & 69.2 & 56.4 & \underline{68.6} & 40.6 & 59.4 & \underline{54.7} & \textbf{69.6} & 39.9 \\
 & RAG 7  & 74.4 & 60.7 & 68.1 & 41.3 & 56.3 & \underline{54.7} & \underline{62.5} & 32.7 \\ \midrule
\multirow{4}{*}{Gemini 1.0 Pro}
 & No RAG & 43.6 & 30.8 & 37.4 & 17.3 & 31.3 & 23.4 & 23.2 & 7.7 \\
 & RAG 3  & 61.5 & 56.4 & 47.8 & 30.0 & 43.8 & 38.5 & 28.6 & 18.8 \\
 & RAG 5  & 64.1 & 51.3 & 46.5 & 27.8 & 37.5 & 33.9 & 28.6 & 19.6 \\
 & RAG 7  & 66.7 & 53.9 & 48.1 & 28.2 & 40.6 & 34.4 & 34.0 & 17.3 \\ \midrule
\multirow{4}{*}{Gemini 1.5 Pro}
 & No RAG & 61.5 & 48.7 & 46.8 & 23.8 & 28.1 & 25.0 & 33.9 & 15.2 \\
 & RAG 3  & 71.8 & 62.4 & 55.6 & 35.2 & 46.9 & 39.1 & 50.0 & 26.2 \\
 & RAG 5  & \underline{76.9} & 57.7 & 57.4 & 36.0 & 46.9 & 40.6 & 39.3 & 30.1 \\
 & RAG 7  & 74.4 & 61.5 & 55.1 & 37.3 & 40.6 & 28.7 & 41.1 & 21.4 \\ 
\bottomrule
\end{tabular}
}
\end{table}

\begin{table}[ht]
\centering
\caption{Performance (\%) of open-source models regarding different answer difficulties.}
\scalebox{0.6}{
\begin{tabular}{cc|cc|cc|cc|cc}
\toprule
\multicolumn{2}{c|}{\multirow{2}{*}{Model}} & \multicolumn{2}{c|}{True/False} & \multicolumn{2}{c|}{Drawing} & \multicolumn{2}{c|}{Calculation} & \multicolumn{2}{c}{Hybrid} \\ \cmidrule(lr){3-4} \cmidrule(lr){5-6} \cmidrule(lr){7-8} \cmidrule(lr){9-10} 
\multicolumn{2}{c|}{}                        & Pass Rate & Accuracy            & Pass Rate & Accuracy         & Pass Rate & Accuracy              & Pass Rate & Accuracy \\ 
\midrule
\multirow{9}{*}{Llama 3} 
 & No Fine-tune       & 43.6 & 33.3 & 28.3 & 10.0  & 15.6 & 12.5 & 26.8 & 8.3  \\
 & Code Only          & \underline{82.1} & \underline{71.8} & 59.2 & 42.0 & 34.4 & 31.3 & 60.7 & \textbf{43.6} \\
 & Code+RAG 3       & \textbf{84.6} & 44.0 & 56.9 & 29.0 & \underline{50.0} & 37.5 & \underline{66.1} & 37.2 \\
 & Code+RAG 5       & 66.7 & 36.8 & 53.5 & 25.4 & 37.5 & 28.1 & 60.7 & 36.3 \\
 & Code+RAG 7       & 66.7 & 37.2 & 50.9 & 24.4 & \underline{50.0} & 35.9 & 64.3 & 39.3 \\
 & Doc+Code           & 82.1 & \textbf{73.1} & \underline{64.4} & \underline{43.7} & 40.6 & 31.8 & \textbf{67.9} & 41.3 \\ \cmidrule(lr){1-10}
\multirow{9}{*}{Deepseek Coder}        
 & No Fine-tune        & 66.7 & 41.5 & 47.8 & 22.1 & \textbf{53.1} & \underline{39.4} & 46.4 & 18.2 \\
 & Code Only          & 71.8 & 61.5 & 60.0 & 41.1 & \underline{50.0} & \textbf{45.3} & 62.5 & 42.1 \\
 & Code+RAG 3      & 71.8 & 48.3 & 57.7 & 32.2 & \textbf{53.1} & \textbf{45.3} & 44.6 & 22.8 \\
 & Code+RAG 5       & 71.8 & 53.9 & 50.7 & 29.3 & 40.6 & 34.4 & 39.3 & 28.6 \\
 & Code+RAG 7       & 74.4 & 54.7 & 50.4 & 28.7 & 37.5 & 34.4 & 48.2 & 31.4 \\
 & Doc+Code           & 79.5 & 68.0 & \textbf{66.2} & \textbf{46.0} & 37.5 & 34.4 & \underline{66.1} & \underline{42.3} \\
\bottomrule
\end{tabular}
}
\end{table}

\begin{table}[ht]
\centering
\caption{Performance (\%) of closed-source models regarding different question types.}
\scalebox{0.68}{
\begin{tabular}{cc|cc|cc}
\toprule
\multicolumn{2}{c|}{\multirow{2}{*}{Model}} & \multicolumn{2}{c|}{Easy} & \multicolumn{2}{c}{Hard} \\
\cmidrule(lr){3-4} \cmidrule(lr){5-6} 
\multicolumn{2}{c|}{} & Pass Rate & Accuracy & Pass Rate & Accuracy         \\ 
\midrule
\multirow{4}{*}{Claude 3 Haiku}
 & No RAG & 55.7 & 31.1 & 29.3 & 10.2 \\
 & RAG 3  & 65.3 & 42.8 & 34.9 & 18.2 \\
 & RAG 5  & 62.7 & 42.0 & 33.0 & 18.4 \\
 & RAG 7  & 64.3 & 48.0 & 30.2 & 19.5 \\ \midrule
\multirow{4}{*}{Claude 3 Sonnet}
 & No RAG & 56.0 & 30.5 & 21.7 & 9.0  \\
 & RAG 3  & 58.7 & 40.4 & 27.8 & 17.0 \\
 & RAG 5  & 61.0 & 42.1 & 25.9 & 15.4 \\
 & RAG 7  & 65.7 & 46.0 & 32.6 & 18.4 \\ \midrule
\multirow{4}{*}{Claude 3 Opus} 
 & No RAG & 68.0 & 44.6 & 41.0 & 20.6 \\
 & RAG 3  & 71.0 & \underline{54.7} & 45.3 & 30.5 \\
 & RAG 5  & 71.7 & 53.7 & 47.6 & \textbf{32.5} \\
 & RAG 7  & \underline{73.3} & \textbf{55.8} & 48.6 & \underline{32.3} \\ \midrule
\multirow{4}{*}{GPT-3.5}        
 & No RAG & 62.3 & 32.2 & 32.6 & 12.4 \\
 & RAG 3  & 66.7 & 42.4 & 40.6 & 15.0 \\
 & RAG 5  & 64.3 & 44.3 & 40.6 & 16.8 \\
 & RAG 7  & 66.7 & 44.2 & 39.2 & 17.1 \\ \midrule
\multirow{4}{*}{GPT-4 turbo}          
 & No RAG & \underline{73.3} & 39.9 & 42.0 & 18.8 \\
 & RAG 3  & 71.3 & 43.6 & 44.8 & 25.7 \\
 & RAG 5  & \textbf{76.0} & 48.8 & 58.5 & 26.6 \\
 & RAG 7  &  \underline{73.3} & 48.6 & 55.2 & 25.6 \\ \midrule
\multirow{4}{*}{GPT-4o}         
 & No RAG & 66.0 & 43.6 & 56.1 & 26.0 \\
 & RAG 3  & 70.7 & 52.4 & \underline{59.9} & 27.6 \\
 & RAG 5  & 71.0 & 51.3 & \textbf{64.2} & 30.2 \\
 & RAG 7  & 72.7 & 54.0 & 59.4 & 26.6 \\ \midrule
\multirow{4}{*}{Gemini 1.0 Pro} 
 & No RAG & 47.3 & 25.1 & 19.8 & 7.2  \\
 & RAG 3  & 61.3 & 44.1 & 25.5 & 13.3 \\
 & RAG 5  & 59.7 & 41.0 & 25.0 & 12.2 \\
 & RAG 7  & 62.3 & 41.9 & 26.4 & 11.6 \\ \midrule
\multirow{4}{*}{Gemini 1.5 Pro} 
 & No RAG & 57.0 & 33.1 & 28.8 & 13.1 \\
 & RAG 3  & 67.3 & 45.7 & 39.2 & 23.7 \\
 & RAG 5  & 69.0 & 47.1 & 38.2 & 23.4 \\
 & RAG 7  & 67.0 & 47.5 & 35.9 & 21.9 \\ 
\bottomrule
\end{tabular}
}
\end{table}

\begin{table}[ht]
\centering
\caption{Performance (\%) of open-source models regarding different question types.}
\scalebox{0.6}{
\begin{tabular}{cc|cc|cc}
\toprule
\multicolumn{2}{c|}{\multirow{2}{*}{Model}} & \multicolumn{2}{c|}{Easy} & \multicolumn{2}{c}{Hard} \\
\cmidrule(lr){3-4} \cmidrule(lr){5-6} 
\multicolumn{2}{c|}{} & Pass Rate & Accuracy & Pass Rate & Accuracy         \\ 
\midrule
\multirow{9}{*}{Llama 3} 
 & No Fine-tune     & 36.3 & 16.2 & 17.5 & 5.4  \\
 & Code Only        & 64.3 & 47.7 & 52.8 & 38.1 \\
 & Code+RAG 3       & 61.7 & 29.6 & 56.6 & 34.5 \\
 & Code+RAG 5       & 56.7 & 25.0 & 50.9 & 31.4 \\
 & Code+RAG 7       & 56.0 & 26.0 & 50.0 & 30.2 \\
 & Doc+Code         & \underline{67.7} & \underline{49.1} & \textbf{60.4} & \underline{39.0} \\ \cmidrule(lr){1-6}
\multirow{9}{*}{Deepseek Coder}        
 & No Fine-tune     & 58.3 & 31.8 & 36.8 & 13.6 \\
 & Code Only        & 64.7 & 46.1 & 54.7 & 38.7 \\
 & Code+RAG 3       & 60.0 & 34.7 & 52.8 & 31.1 \\
 & Code+RAG 5       & 55.0 & 31.5 & 43.9 & 31.2 \\
 & Code+RAG 7       & 54.7 & 32.9 & 46.2 & 29.2 \\
 & Doc+Code         & \textbf{69.7} & \textbf{50.6} & \underline{59.4} & \textbf{40.8} \\
\bottomrule
\end{tabular}
}
\end{table}

\clearpage
\newpage

\section{Experimental Results of Best-performing Models}
\label{sec:experimental-results-of-best-performing-models}

\begin{figure}[H]
    \centering
    \begin{subfigure}[b]{0.48\textwidth}
        \centering
        \includegraphics[width=\textwidth]{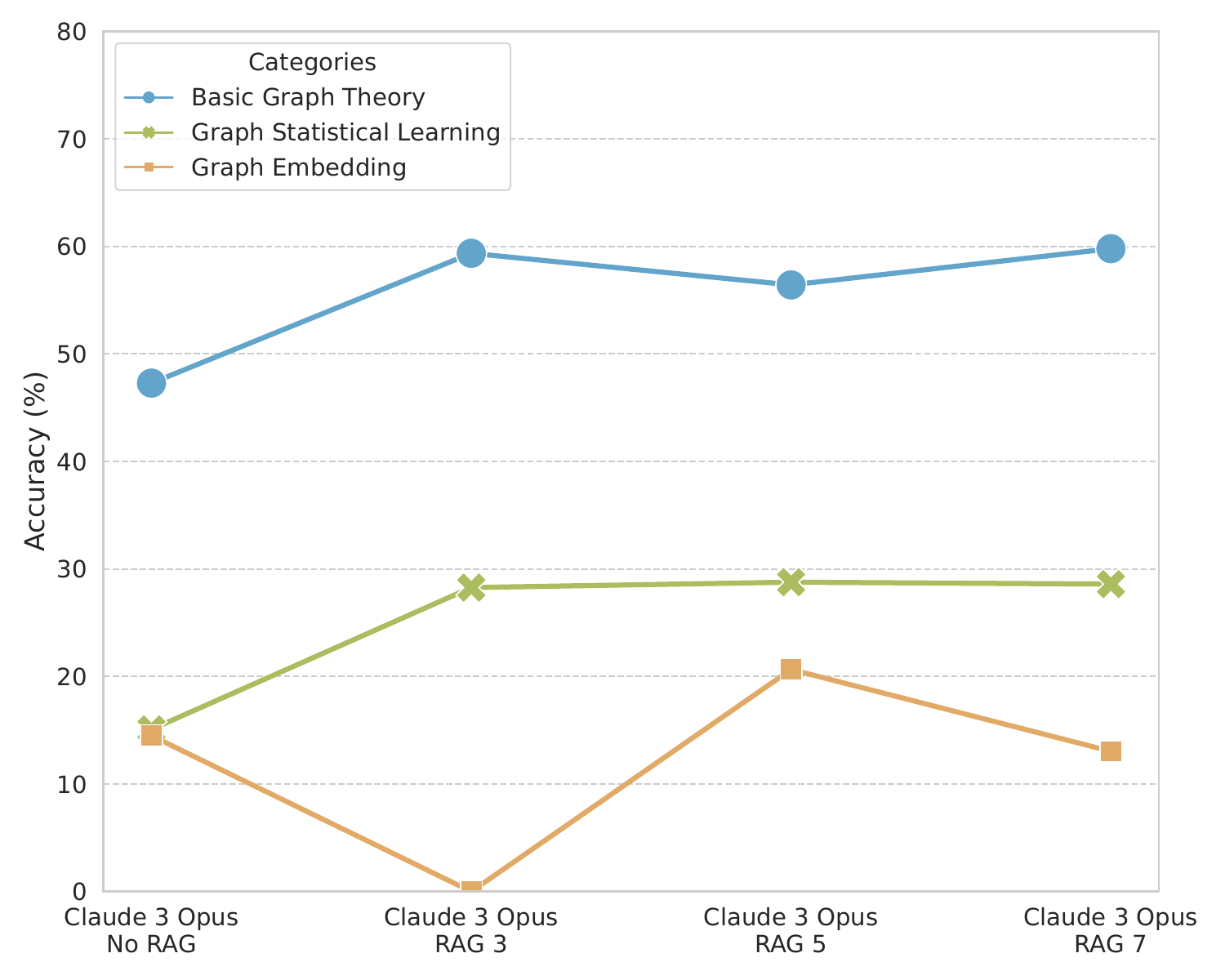}
        \label{fig:close-source-pass-rate-category}
    \end{subfigure}
    \hfill
    \begin{subfigure}[b]{0.48\textwidth}
        \centering
        \includegraphics[width=\textwidth]{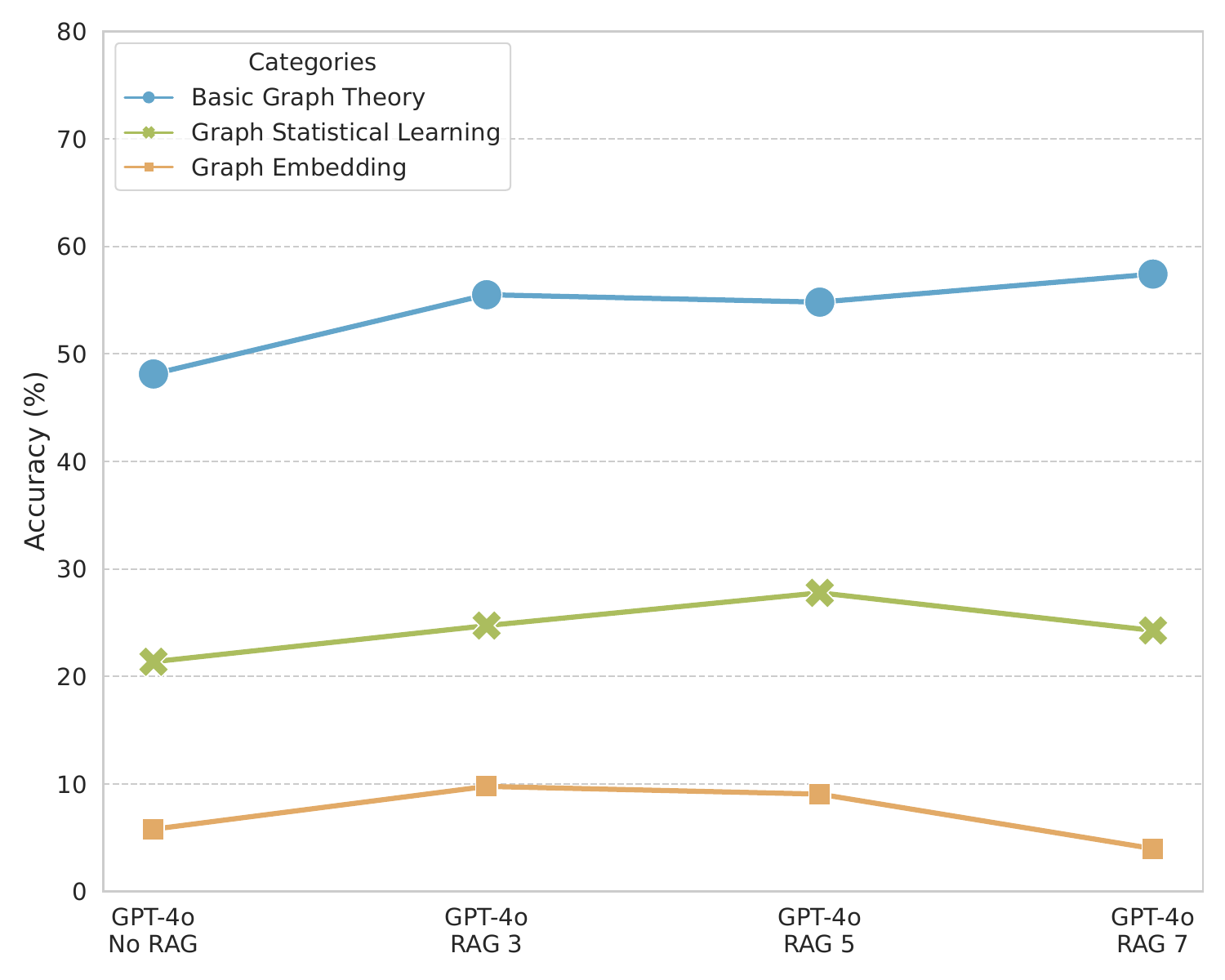}
        \label{fig:gpt4o-accuracy-category}
    \end{subfigure}
    \caption{Accuracy of closed-source models regarding different task categories.}
    \label{fig:close-source-models-pass-rate-and-accuracy-category}
\end{figure}

\begin{figure}[H]
    \centering
    \begin{subfigure}[b]{0.48\textwidth}
        \centering
        \includegraphics[width=\textwidth]{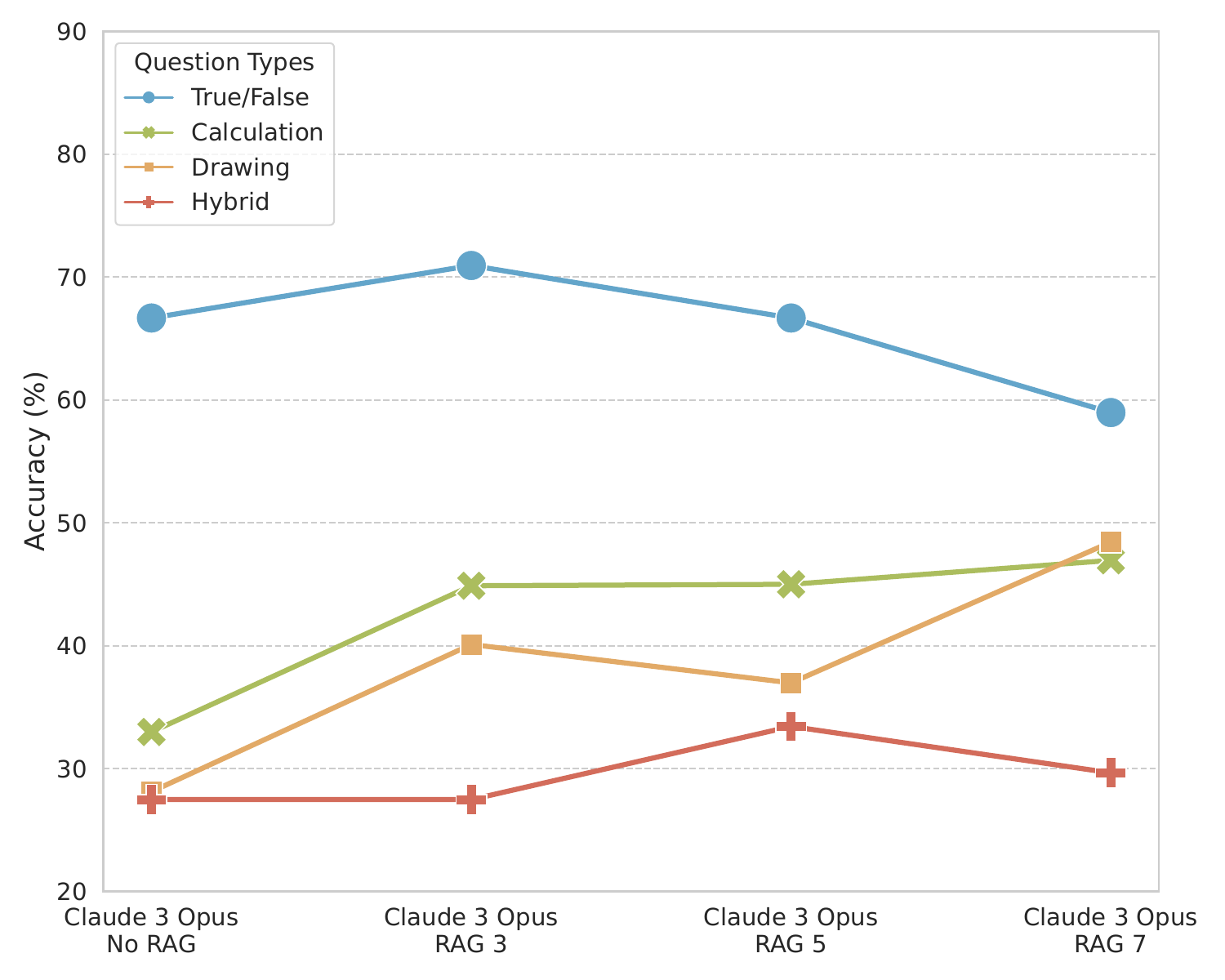}
        \label{fig:close-source-pass-rate-question-type}
    \end{subfigure}
    \hfill
    \begin{subfigure}[b]{0.48\textwidth}
        \centering
        \includegraphics[width=\textwidth]{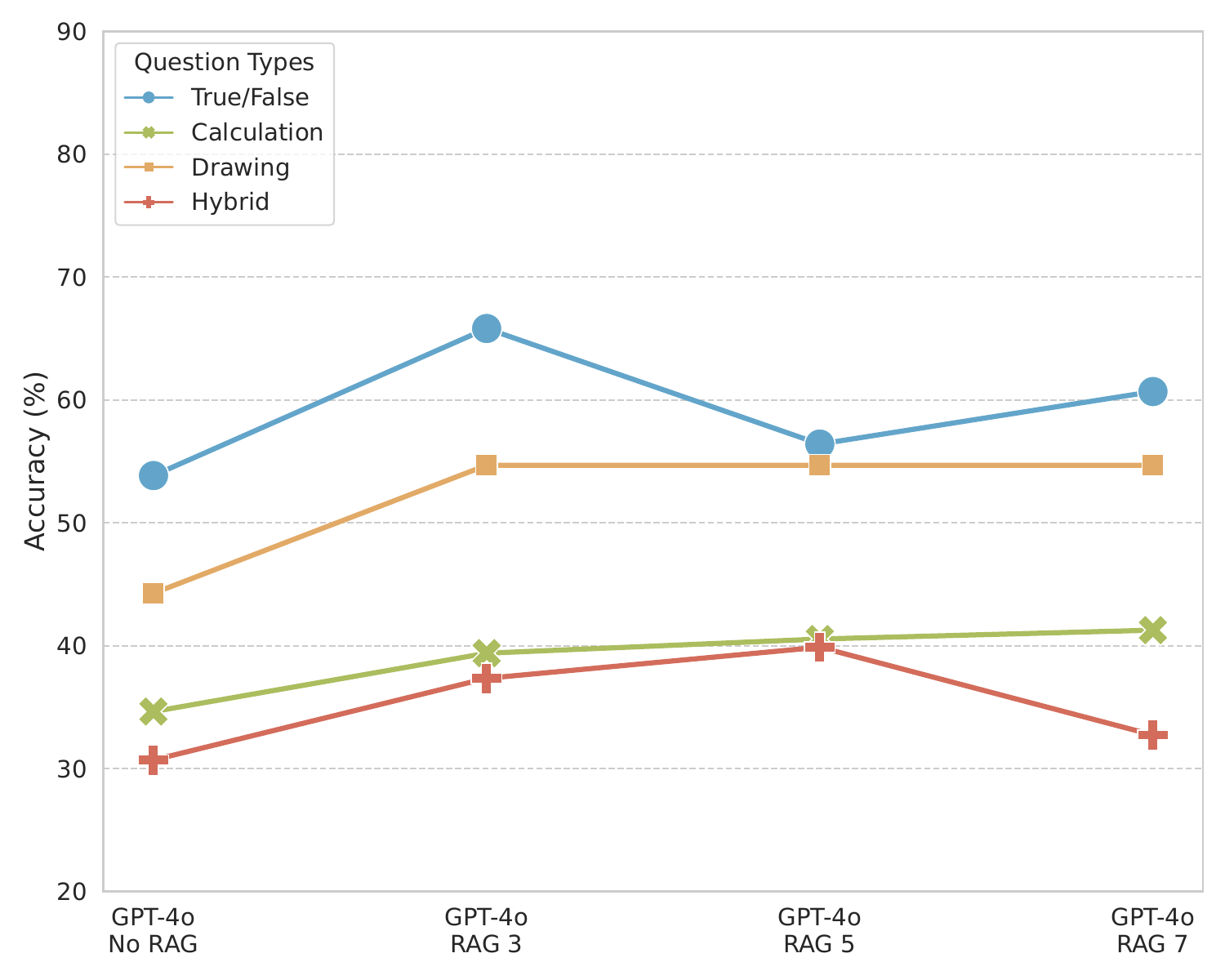}
        \label{fig:gpt4o-accuracy-question-type}
    \end{subfigure}
    \caption{Accuracy of closed-source models regarding different answer difficulties.}
    \label{fig:close-source-models-pass-rate-and-accuracy-question-type}
\end{figure}

\begin{figure}[H]
    \centering
    \begin{subfigure}[b]{0.48\textwidth}
        \centering
        \includegraphics[width=\textwidth]{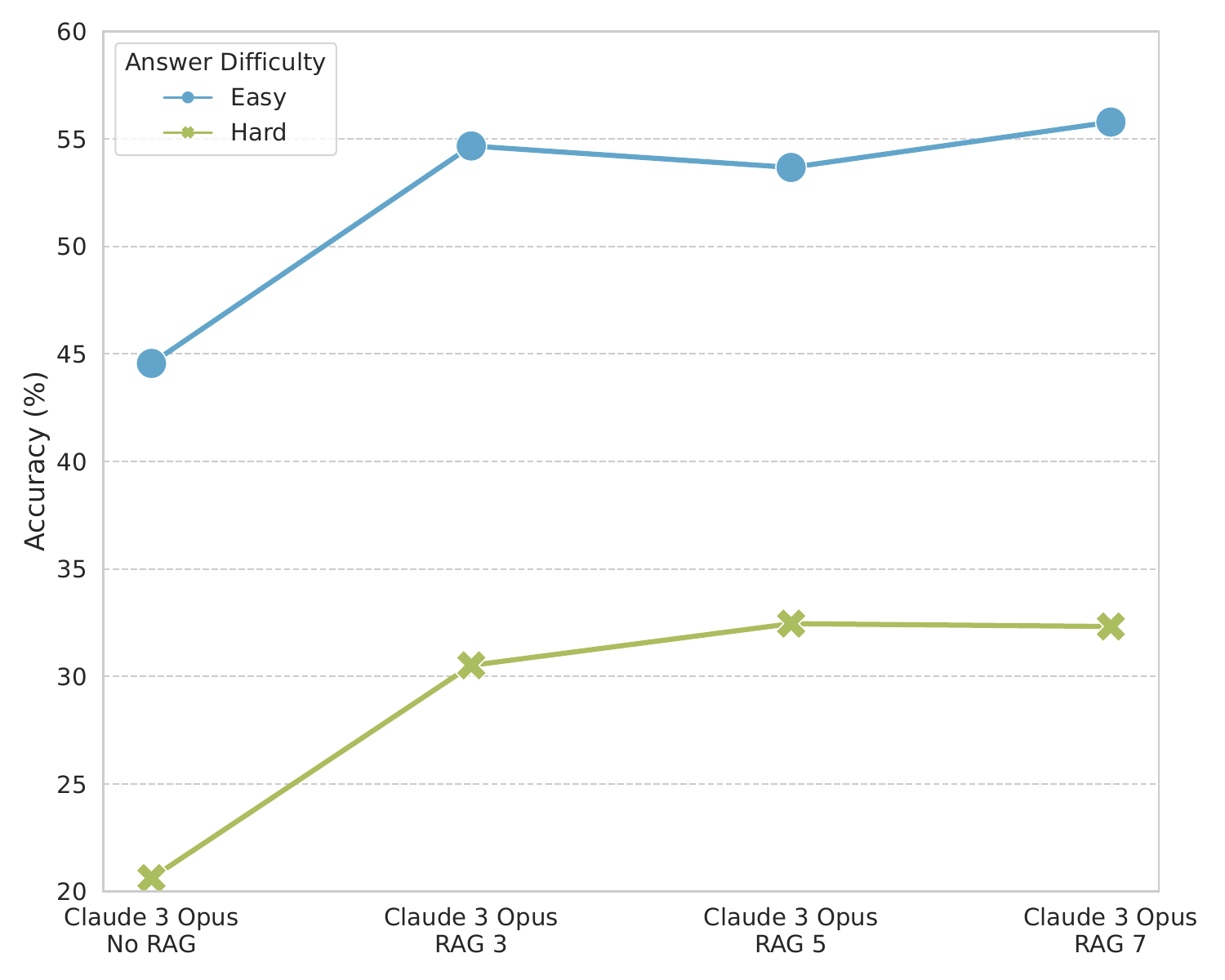}
        \label{fig:close-source-pass-rate-answer-difficulty}
    \end{subfigure}
    \hfill
    \begin{subfigure}[b]{0.48\textwidth}
        \centering
        \includegraphics[width=\textwidth]{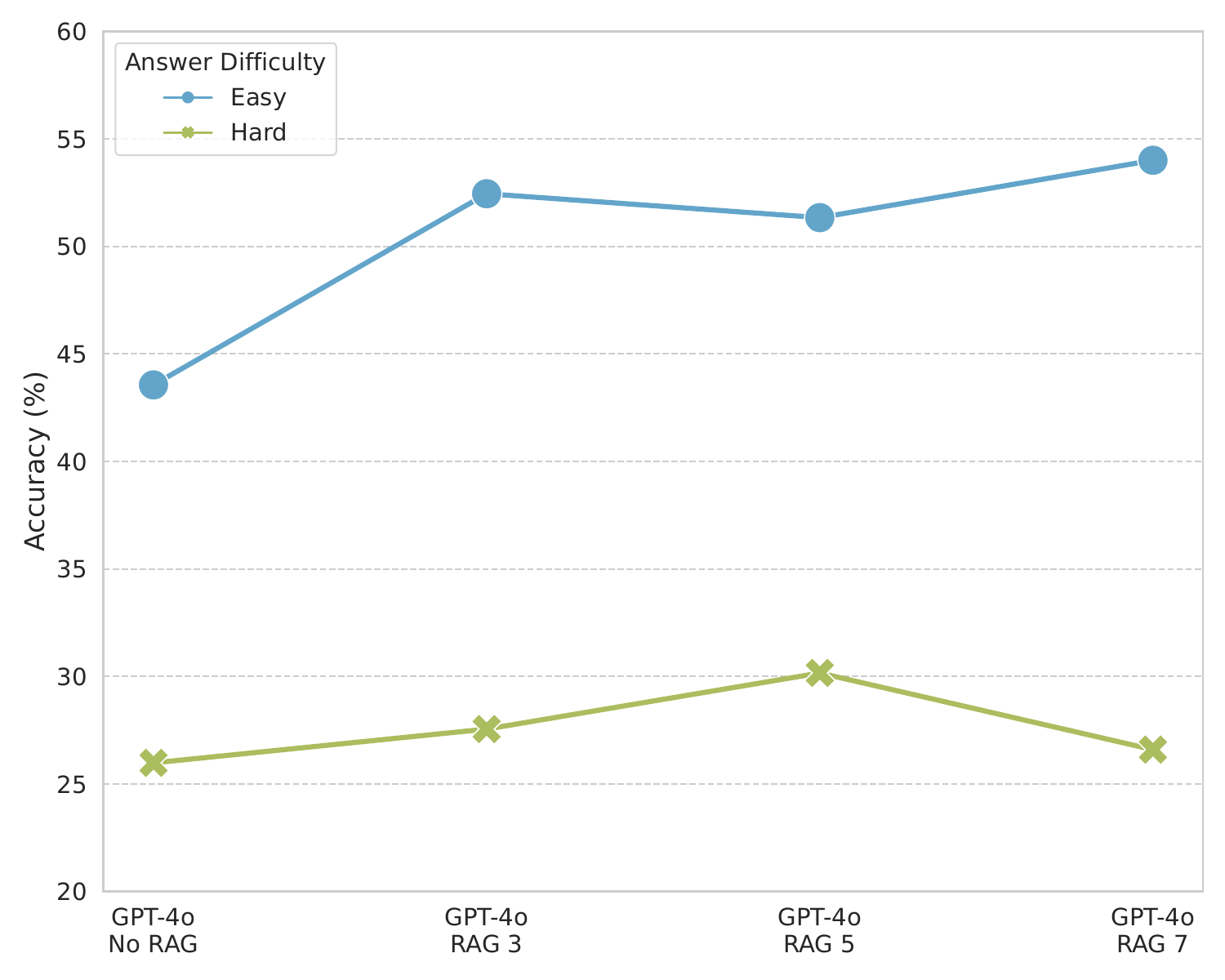}
        \label{fig:gpt-4o-accuracy-answer-difficulty}
    \end{subfigure}
    \caption{Accuracy of closed-source models regarding different question types.}
    \label{fig:close-source-models-pass-rate-and-accuracy-answer-difficulty}
\end{figure}

\clearpage

\begin{figure}[t]
    \centering
    \begin{subfigure}[b]{0.48\textwidth}
        \centering
        \includegraphics[width=\textwidth]{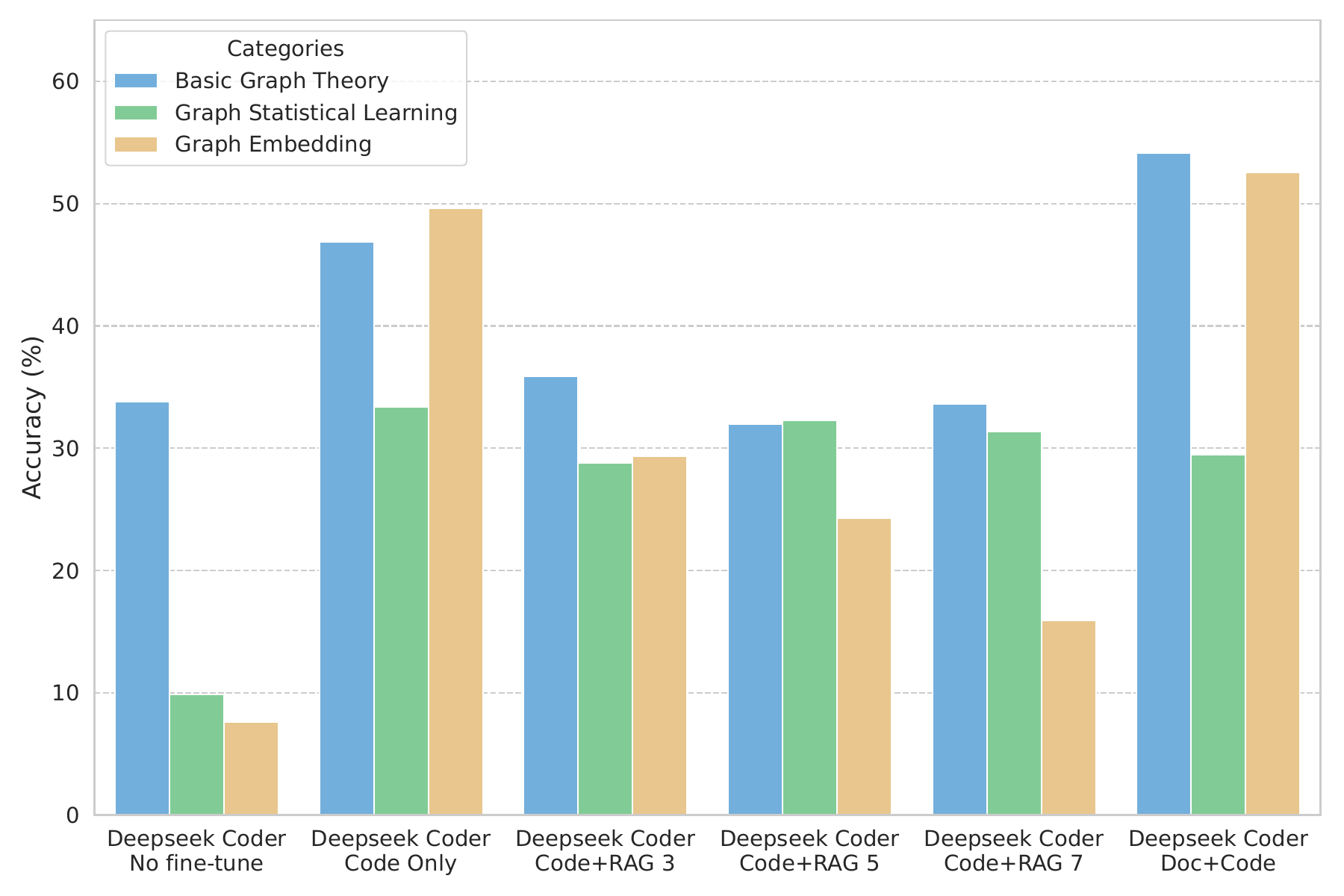}
        \label{fig:open-source-pass-rate-category}
    \end{subfigure}
    \hfill
    \begin{subfigure}[b]{0.48\textwidth}
        \centering
        \includegraphics[width=\textwidth]{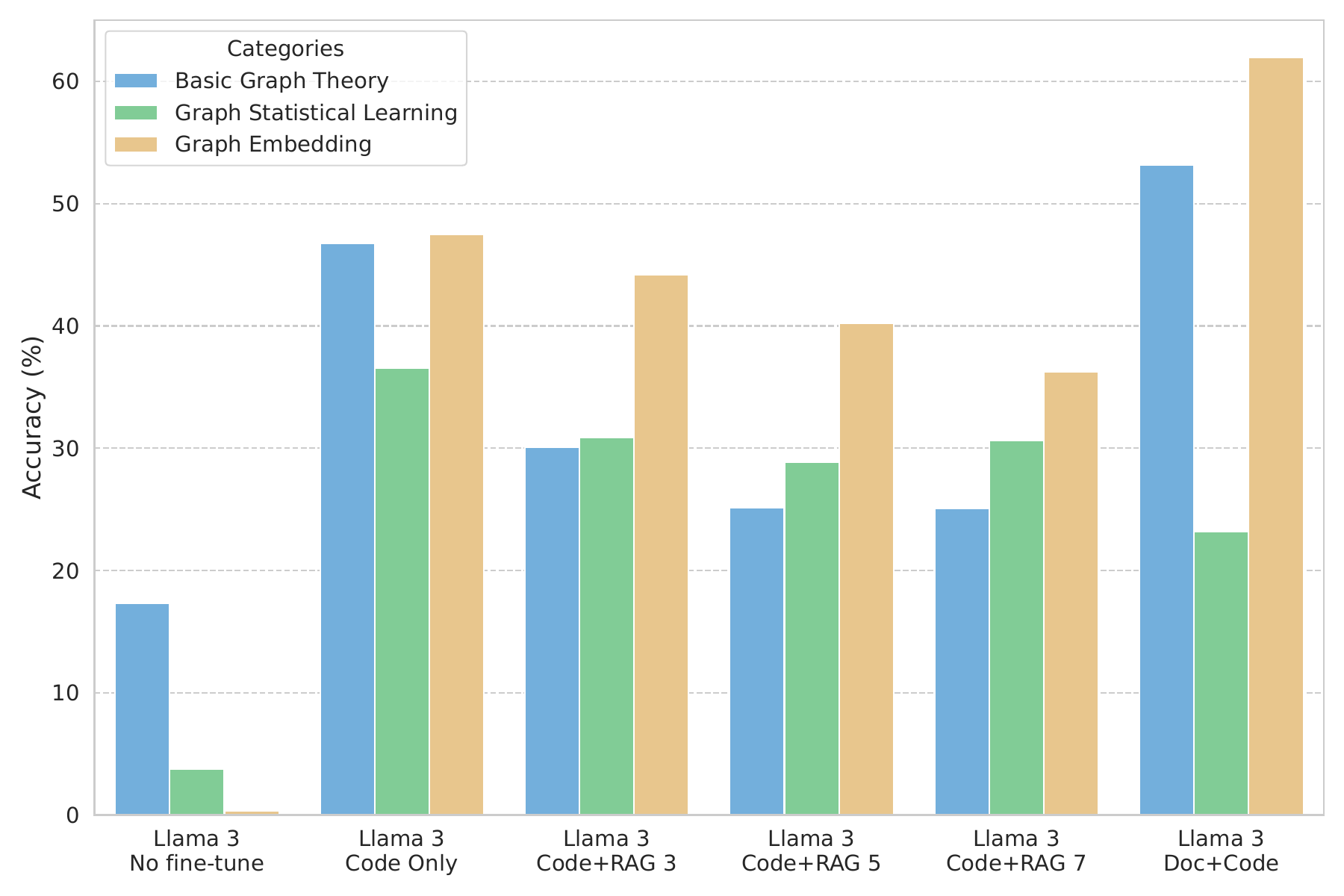}
        \label{fig:llama3-accuracy-category}
    \end{subfigure}
    \caption{Accuracy of open-source models regarding different task categories.}
    \label{fig:open-source-models-pass-rate-and-accuracy-category}
\end{figure}

\begin{figure}[t]
    \centering
    \begin{subfigure}[b]{0.48\textwidth}
        \centering
        \includegraphics[width=\textwidth]{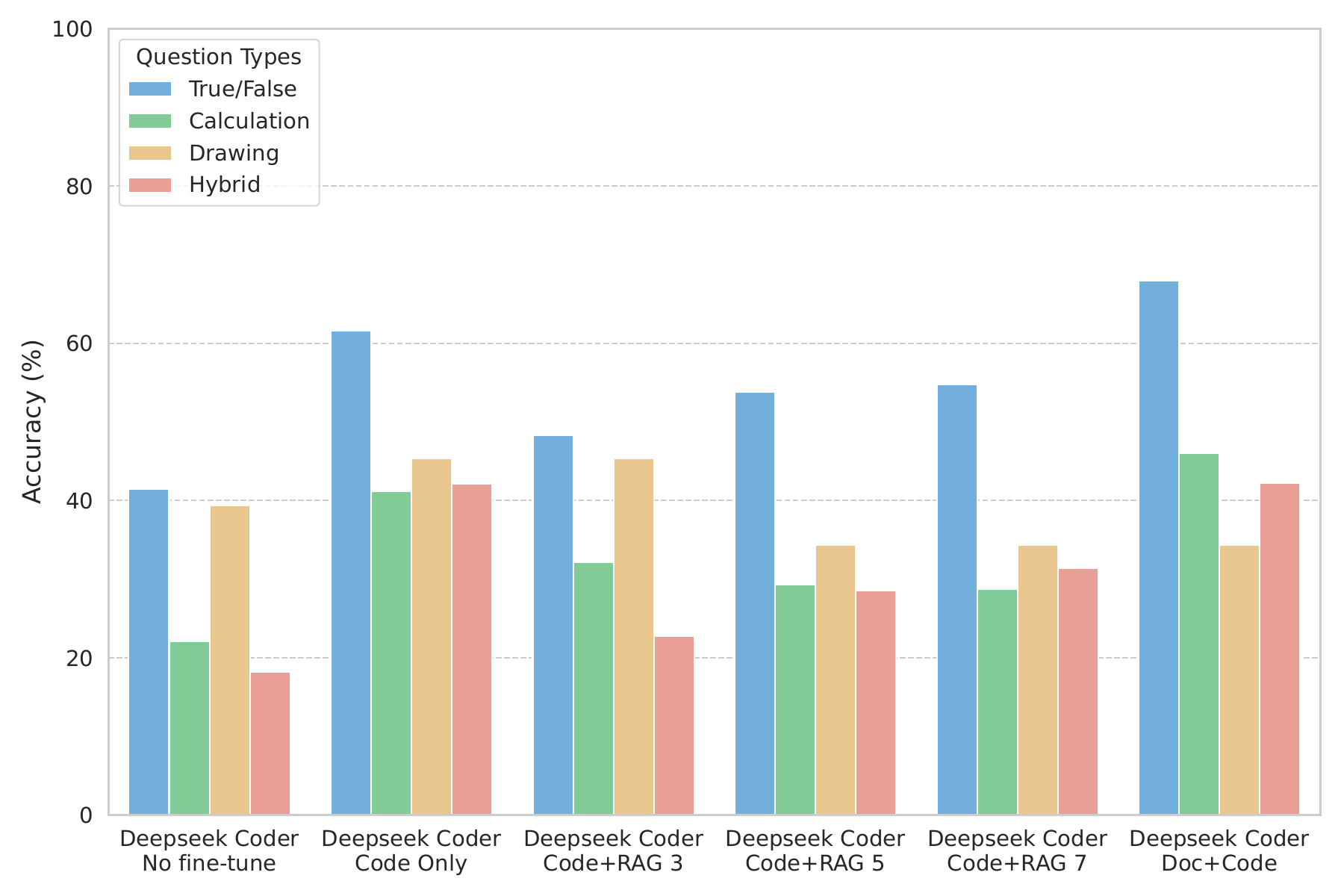}
        \label{fig:open-source-pass-rate-question-type}
    \end{subfigure}
    \hfill
    \begin{subfigure}[b]{0.48\textwidth}
        \centering
        \includegraphics[width=\textwidth]{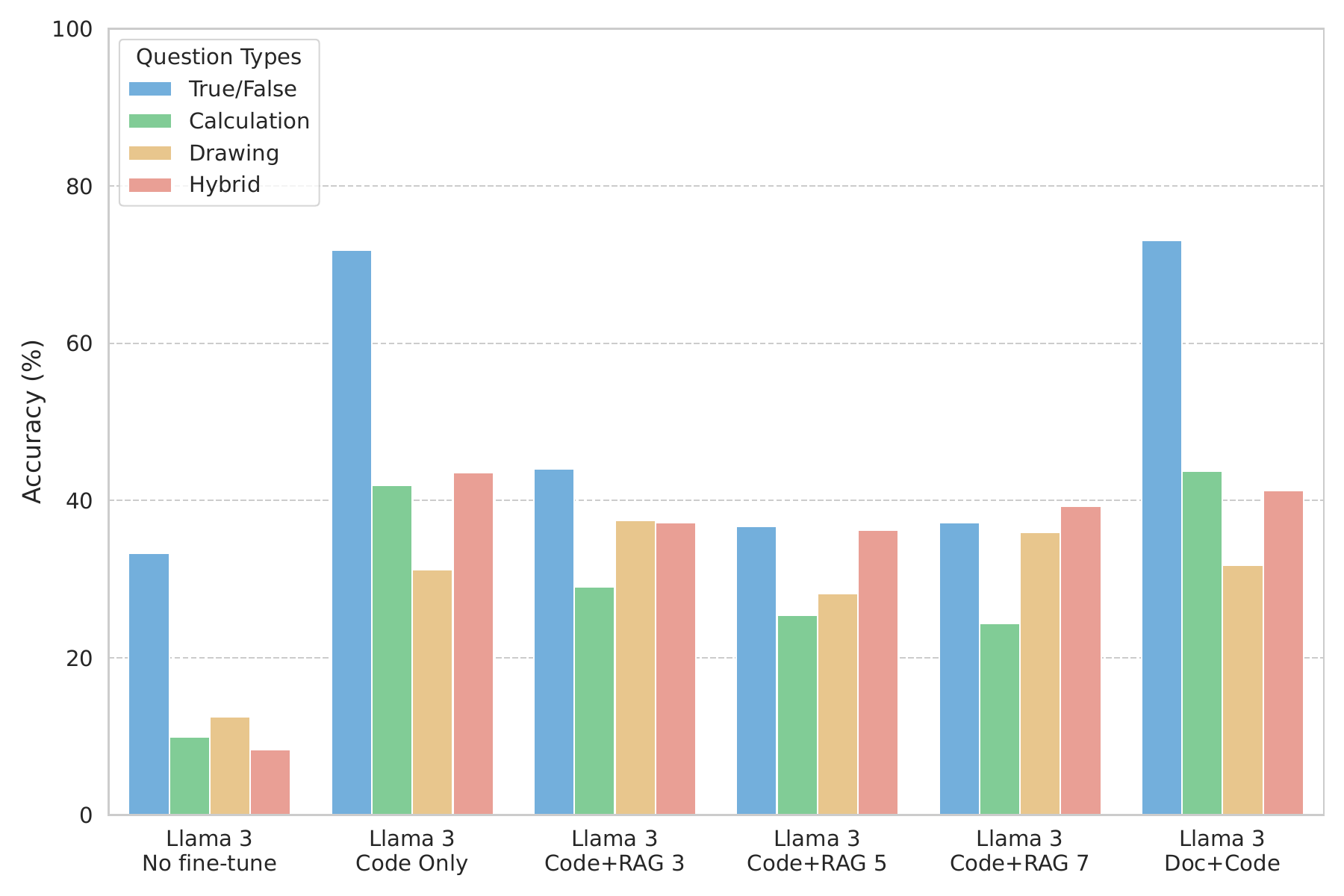}
        \label{fig:llama3-accuracy-type}
    \end{subfigure}
    \caption{Accuracy of open-source models regarding different answer difficulties.}
    \label{fig:open-source-models-pass-rate-and-accuracy-question-type}
\end{figure}

\begin{figure}[t]
    \centering
    \begin{subfigure}[b]{0.48\textwidth}
        \centering
        \includegraphics[width=\textwidth]{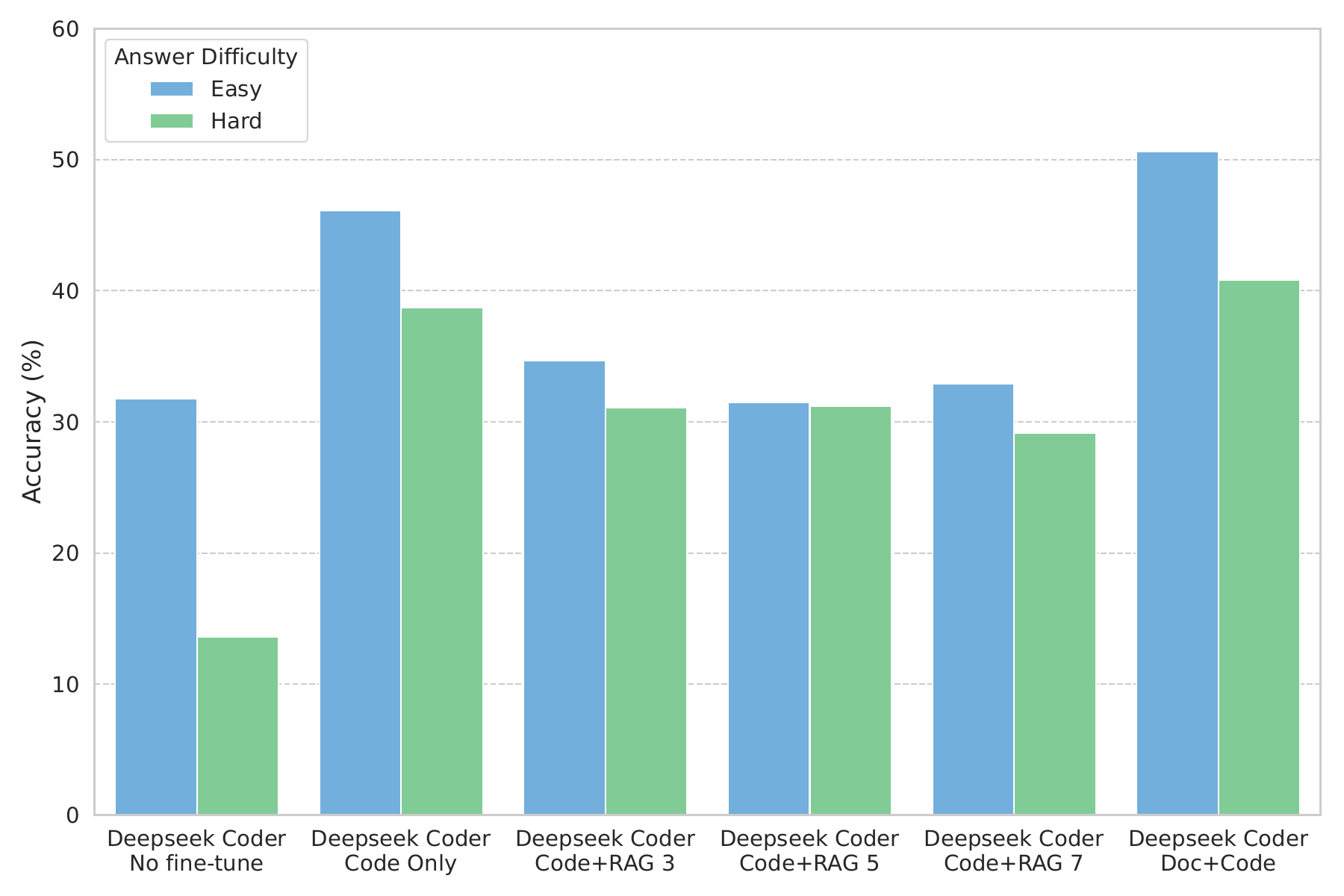}
        \label{fig:open-source-pass-rate-answer-difficulty}
    \end{subfigure}
    \hfill
    \begin{subfigure}[b]{0.48\textwidth}
        \centering
        \includegraphics[width=\textwidth]{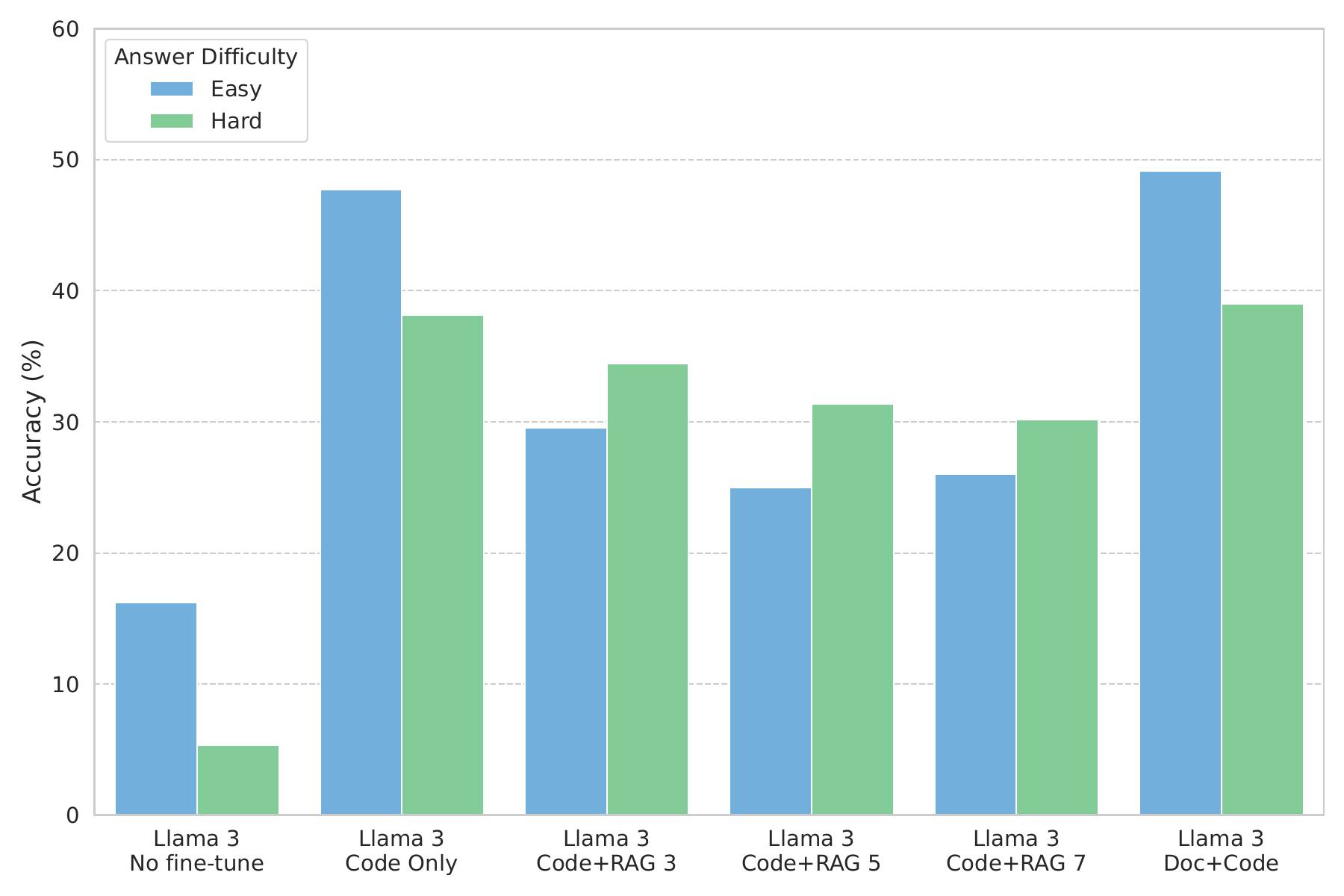}
        \label{fig:llama3-accuracy-answer-difficulty}
    \end{subfigure}
    \caption{Accuracy of open-source models regarding different question types.}
    \label{fig:open-source-models-pass-rate-and-accuracy-answer-difficulty}
\end{figure}

\clearpage
\section{Annotation Manual for ProGraph Benchmark}
\label{annotation-manual}

\textbf{Annotation}
This manual provides detailed guidelines for annotating a benchmark dataset by creating and documenting tasks related to graph-related Python packages such as NetworkX and graph. Each annotator will document a function, propose a question, create a graph, and provide a reference code with execution results.

\begin{enumerate}
    \item \textbf{Objective}
    \begin{itemize}
        \item To create a high-quality benchmark dataset for evaluating the performance of graph-related functions in Python packages.
        \item To document function usage, practical questions, graph creation, corresponding results, and key APIs.
    \end{itemize}

    \item \textbf{Annotation Tools}
    \begin{itemize}
        \item \textbf{Graph Libraries:} NetworkX, igraph, CDlib, graspologic, Karate Club, Little Ball of Fur
        \item \textbf{Code Editor:} Any Python IDE (e.g., PyCharm, VS Code, Jupyter Notebook)
        \item \textbf{Documentation Resources:} Official documentation for the 6 Python libraries.
    \end{itemize}

    \item \textbf{Data Types}
    \begin{itemize}
        \item \textbf{Function Documentation:} Description and usage of a specific function.
        \item \textbf{Graph:} Manually crafted or randomly generated graphs.
        \item \textbf{Code:} Python code snippets for graph creation and function execution.
        \item \textbf{Results:} Output from executing the provided code.
        \item \textbf{Key APIs:} APIs that are crucial and indispensable for solving a problem.
    \end{itemize}
    \item \textbf{Annotation Guidelines}
    \begin{itemize}
        \item \textbf{Function Documentation:}
        \begin{itemize}
            \item Select an API from the provided graph-related packages.
            \item Document the API name, parameters, return type, and a brief description.
        \end{itemize}
        \item \textbf{Proposing a Question:}
        \begin{itemize}
            \item Formulate a clear, practical question that the API can solve.
            \item Ensure the question is specific and relevant to the API's capabilities.
        \end{itemize}
        \item \textbf{Creating a Graph:}
        \begin{itemize}
            \item \textbf{Manual Crafting:} Draw a graph that fits the proposed question.
            \item \textbf{Random Generation:} Use code to generate a random graph appropriate for the function.
            \item \textbf{Graph Description:} Provide a brief description of the graph structure.
        \end{itemize}
        \item \textbf{Reference Code:}
        \begin{itemize}
            \item Write Python code to create the graph.
            \item Include the API call with appropriate parameters.
            \item Ensure the code is well-commented and readable.
        \end{itemize}
        \item \textbf{Execution Result:}
        \begin{itemize}
            \item Execute the code and record the output.
            \item Provide a detailed explanation of the result.
        \end{itemize}
    \end{itemize}

    \item \textbf{Annotation Process}
    \begin{itemize}
        \item \textbf{Step-by-Step Procedure:}
        \begin{enumerate}
            \item \textbf{API Selection:} Choose an API from the provided documentation.
            \item \textbf{Question Formulation:} Develop a practical question for the API.
            \item \textbf{Graph Creation:} Create a graph manually or using code.
            \item \textbf{Code Writing:} Write reference code to demonstrate the API.
            \item \textbf{Result Recording:} Execute the code and document the output.
            \item \textbf{Review and Submission:} Review the annotation for accuracy and clarity, then submit.
        \end{enumerate}
        \item \textbf{Examples:}
        \begin{itemize}
            \item Provide examples for each step to guide annotators.
        \end{itemize}
    \end{itemize}

    \item \textbf{Quality Control}
    \begin{itemize}
        \item \textbf{Review Process:}
        \begin{itemize}
            \item Conduct peer reviews of annotations to ensure consistency and accuracy.
            \item Provide feedback and request revisions if necessary.
        \end{itemize}
        \item \textbf{Consistency Checks:}
        \begin{itemize}
            \item Ensure all annotations follow the same structure and guidelines.
            \item Verify the correctness of code and results.
        \end{itemize}
    \end{itemize}

    \item \textbf{Common Issues}
    \begin{itemize}
        \item \textbf{Ambiguity in Questions:} Ensure questions are specific and clear.
        \item \textbf{Errors in Code:} Double-check code for syntax and logical errors.
        \item \textbf{Inconsistent Results:} Verify that results match the expected output.
    \end{itemize}

    \item \textbf{Annotation Tips}
    \begin{itemize}
        \item Be precise and consistent in the document.
        \item Use the provided examples as references.
        \item Clarify any doubts with the project coordinator.
    \end{itemize}

    \item \textbf{Frequently Asked Questions (FAQs)}
    \begin{itemize}
        \item \textbf{Q:} What if I encounter an error in the API execution?
        \item \textbf{A:} Check the document and debug the code. If the error persists, seek help from the coordinator.
        \item \textbf{Q:} How detailed should the graph description be?
        \item \textbf{A:} Provide enough detail to understand the graph structure and its relevance to the API.
    \end{itemize}

\end{enumerate}

\end{document}